%% file: main.tex
\pdfoutput=1
\documentclass[11pt,letterpaper]{article}
\usepackage[natbibapa]{apacite}
\usepackage[margin=1in]{geometry}
\usepackage{times}
\usepackage[T1]{fontenc}
\usepackage[utf8]{inputenc}

\usepackage{hyperref}
\usepackage{url}
\usepackage{amsmath,amssymb}
\usepackage{booktabs}
\usepackage{graphicx}
\usepackage{multirow}
\usepackage{array}
\usepackage{xcolor}
\usepackage{microtype}
\usepackage{tikz}
\usepackage{seqsplit}
\usepackage{amsthm}
\usepackage{float}
\usepackage{caption}
\usetikzlibrary{arrows.meta,positioning,calc,fit}
\newtheorem{lemma}{Lemma}
\hypersetup{colorlinks=true,linkcolor=black,citecolor=black,urlcolor=blue}

\newcommand{\Sone}{\text{\textup{SAS}}}
\newcommand{\Mone}{\text{\textup{MAS}}}
\newcommand{\Bx}{B_{1\times}}
\newcommand{\passone}{pass$^{1}$}

\title{Does Multi-Agent Collaboration Pay When Agents Serve Users?\\ Testing Agent Scaling in Dual-Control Settings}

\author{Ethan Yang \qquad Xiangcheng Shi\\[3pt]
University of Pennsylvania\\
\texttt{\{eyyang, xcshi\}@seas.upenn.edu}}
\date{October 2026}

\begin{document}
\shortcites{kim2026capable,barres2025tau2,yao2024tau,tran2026single,jwalapuram2026illusion,wang2024token,kapoor2024agents,kapoor2025hal,gao2025single,yang2025debate,li2026scaling,yu2026bottleneck,seshadri2026lost,zhou2026sim2real,cuadron2025saber,kang2026policyguard,fourney2024magentic,cemri2025mast,schaeffer2023mirage,hofman2023prereg}

\maketitle

\input{sections/00_abstract}
\input{sections/01_introduction}
\input{sections/02_related}
\input{sections/03_design}
\input{sections/04_results}
\input{sections/05_discussion}
\input{sections/06_conclusion}

\bibliography{references}
\bibliographystyle{apacite}

\appendix
\input{sections/appendix}

\end{document}

%% file: sections/00_abstract.tex
\begin{abstract}
Multi-agent systems (MAS) built from language models are increasingly deployed on the premise that collaboration among agents outperforms a single-agent system (SAS), yet when collaboration pays remains poorly understood. The most comprehensive study to date answers with an empirical capability-saturation rule, offered as a practical selection rule rather than a universal law: coordination helps while the underlying SAS is weak and hurts once it is strong, with the crossing near 45\% single-agent accuracy. That rule was derived on autonomous benchmarks, where no user takes part. We test it where agents increasingly operate: in conversation with a user under a written policy, on $\tau^2$-bench, the dual-control benchmark, where the user supplies information turn by turn and, in its telecom domain, acts on the shared environment as well. On its airline, retail and telecom domains we compare a SAS with a centralized MAS built from the same model (an orchestrator, a database agent and a policy advisor) for six models from three families (four, from two families, in the primary analysis), over about 7,950 simulated conversations, with each system held to one completion-token ledger per episode. The study was pre-registered with a public change log; the MAS is reported as built, which leaves one pre-registered validity bar unmet and labelled as a deviation (its planner never saw the tool schemas). Under a fixed simulated user no cell shows a MAS benefit: the fitted gain is negative across the observed capability range, with a weighted mean of $-28$ points (95\% CI $[-44, -12]$) over the 12 cells of the primary fit, and the MAS is worse in 12 of the 13 cells that meet the mechanical validity bars and tied in the thirteenth, a cell near 45\% where the benchmark's default simulator yields a 10-point MAS gain. On the two cells where the principle predicts a benefit, the MAS loses by 13 and 22 points, and removing the budget ceiling brings it at best level with the SAS at 5.5 times its median token spend. The design could not resolve whether the gain declines with capability: a capability index measured on held-out domains gives a slope of $-0.41$ with a 95\% CI of $[-1.51, +0.68]$, because model-by-domain effects dominate the variation between cells. The mechanism that would justify a MAS here is not detectable: the MAS shows no measurable reduction in wrongful database writes and converts other failures into budget exhaustion and inaction. For agents that serve people, these results make the SAS the default for this architecture and leave multi-agent designs to show compute-matched evidence from the setting itself. We release all trajectories, budget ledgers, prompts and code.
\end{abstract}

%% file: sections/01_introduction.tex
\section{Introduction}
\label{sec:intro}

Language-model agents are moving from research prototypes into deployed products, and one design question now precedes almost every deployment: should a task be handled by a single-agent system (SAS) or by a multi-agent system (MAS) whose agents plan, check and specialize? Multi-agent frameworks have multiplied, and industrial reports describe large gains from orchestrating several agents, at many times the tokens of a single one \citep{anthropic2025multiagent}. Studies that hold compute fixed \citep{tran2026single} or account for cost after the fact \citep{jwalapuram2026illusion} often find the SAS as strong or stronger. The choice is consequential, since a MAS multiplies model calls, latency and points of failure.

\citet{kim2026capable} gave the most systematic answer to date, and with it the beginnings of a science of agent scaling. Across 260 configurations spanning six benchmarks, five architectures and nine models, the capability of the underlying SAS was the strongest predictor of what coordination adds: a MAS pays while the SAS is weak and costs once it is strong, with the fitted crossing near 45\% single-agent accuracy. In a validation on 16 model-by-benchmark cells the rule predicted in 15 the direction of the gain of the best of its four MAS architectures over the SAS. The authors offer it as a practical selection rule derived on the tasks and architectures they tested rather than a universal scaling law; we call it the capability-saturation principle for short, since what we test is whether the rule, as practitioners would apply it, carries over to a new setting.

Every benchmark behind this principle is autonomous: the agent receives a task and acts alone. Many agents now reaching users operate in a different regime. The agent converses with a person who supplies information late, changes goals and asks for what the policy forbids, and in the dual-control case of \citet{barres2025tau2} the person holds tools of their own and acts on the shared environment too. This regime bears on a MAS in three ways. The conversation has one user-facing locus, so most generic MAS architectures have no deployable analogue. Input tokens dominate the cost, so what is held constant when comparing systems decides the outcome. And the user is simulated by a model whose choice alone can move scores as much as the per-cell effects reported for the principle \citep{seshadri2026lost,zhou2026sim2real}. Whether principles derived without a user hold once a user takes part is of both scientific and practical weight.

We present such a test on $\tau^2$-bench \citep{barres2025tau2}, the standard dual-control benchmark, where the reward is the state of a database or device when the conversation ends; its airline and retail domains give tools to the agent alone, and its telecom domain gives the user tools as well. We compare the benchmark's stock SAS with a centralized MAS built from the same model: an orchestrator that plans and talks to the customer, a database agent that executes its reads and writes, and a policy advisor (Figure~\ref{fig:design}). Six models from three families on three domains give 18 cells, of which 16 were run on the registered configuration, for about 7,950 simulated conversations in all. The study was pre-registered, and every change to the plan is dated in a public change log.\footnote{Code, prompts, run configuration, the pre-registered plan with its change log and every derived table are at \url{https://github.com/ethanyyang/agent-scaling-dual-control}; the trajectories with per-call token records, budget ledgers and calibration files are archived at \url{https://doi.org/10.5281/zenodo.23458286}.} Within a cell both systems run the same 50 tasks with three seeds, the MAS under a per-episode completion-token ceiling calibrated from the SAS's own spend. We read the result in one of three ways: the principle \emph{transfers} if the gain declines with capability and crosses zero near 45\%, \emph{shifts} if it crosses elsewhere, and \emph{disappears} if no dependence is detected. Mechanical validity checks separate a MAS that does not help from one that is broken, and the primary test uses a capability index measured on held-out domains, which removes the negative slope that a gain regressed on its own baseline shows by construction.

We find no support for the principle's distinctive prediction in this setting, for this architecture. Under a fixed simulated user no cell shows a MAS benefit: the fitted gain is negative across the observed capability range, with a weighted mean of $-28$ points, and the MAS is worse than the SAS in 12 of the 13 cells that meet the mechanical validity bars and tied in the thirteenth, the cell nearest 45\%, where only a change of simulator produces a 10-point gain (\S\ref{sec:robust}). On the two cells where the principle predicts a benefit, the MAS loses by 13 and 22 points, and with no budget ceiling at all it at best draws level at 5.5 times the SAS's median token spend (\S\ref{sec:transfer}). Whether the gain declines with capability the design could not resolve: the slope test detects no dependence, and its interval cannot exclude a steep decline, because model-by-domain effects dominate the variation between cells (\S\ref{sec:h1}). The mechanism that motivates a MAS in this setting, intercepting wrongful database writes, is not detectable: the MAS does not measurably reduce them and converts other failures into budget exhaustion and inaction (\S\ref{sec:h4}). One qualification runs through the paper. The MAS as built did not meet one of our pre-registered validity bars: its orchestrator planned without the tool schemas that only the database agent holds, and about three quarters of its write orders were not executed exactly as written. The pre-registered response, a fix and a re-run, was built and smoke-tested but not run on the grid; we report the architecture as built and label the deviation (\S\ref{sec:compliance}).

\paragraph{Contributions.} (1) The first test of the capability-saturation principle in the regime where agents serve users in conversation, on the dual-control benchmark. (2) A compute-matching protocol for multi-turn agent systems: one per-episode completion-token ledger for all agents of a system, calibrated per cell. (3) An analysis, supported by a lemma, that separates capability dependence from the coupling between a gain and its own baseline. (4) Evidence on why the MAS fails: a taxonomy separating write-type from word-type violations, per-cell validity checks, and a no-ceiling arm that rules out the budget as the sole cause. (5) Release of all trajectories, ledgers, prompts and code (footnote~1).

%% file: sections/02_related.tex
\section{Background and related work}
\label{sec:related}

\paragraph{The capability-saturation principle.} \citet{kim2026capable} compared a SAS with four generic MAS architectures (independent, centralized, decentralized, hybrid) across six autonomous benchmarks and three model families at three tiers, holding prompts, tools and a per-system reasoning-token budget constant. Their regression explains system performance, not the gain over the SAS, from standardized predictors. The SAS baseline is its strongest positive term (coefficient 0.250) and the interaction of the baseline with the number of agents is negative ($-0.236$); adding agents therefore stops paying once the baseline passes the point at which the agent-count main effect (0.040) and that interaction cancel, $P^{\ast}_{\mathrm{SA}} = -\hat\beta_{\text{agents}}/\hat\beta_{\text{interaction}}$, about 45\% after de-standardisation. The interaction does not survive the authors' cluster-robust correction, and they present the threshold as a practical selection rule rather than a universal scaling principle. Its sign validation (15 of 16 cells on SWE-bench Verified and Terminal-Bench, with the gain taken as the best of the four MAS architectures against the SAS) used 20 tasks per cell, with per-cell intervals of about 20 points. The threshold is global, with no benchmark-specific parameter, so a new setting needs only its own SAS baselines, and the $\tau$-bench family is named in that paper as untested.

\paragraph{Compute-matched comparisons.} Agent evaluations rarely report cost \citep{kapoor2024agents,kapoor2025hal}, and positive results for MAS are seldom compute-matched. Under a shared token budget self-consistency often beats debate and reflection \citep{wang2024token}; under a fixed thinking budget a SAS with perfect context utilisation is in theory at least as information-efficient as any multi-agent decomposition, and in experiment it was best or tied at every budget but the smallest \citep{tran2026single}; automatically generated MAS mostly underperform a five-sample self-consistency baseline at up to ten times its cost, with HLE-Math the exception \citep{jwalapuram2026illusion}. Several results agree with the principle, with the benefit of a MAS shrinking as models improve \citep{gao2025single,yang2025debate,yu2026bottleneck}; \citet{li2026scaling} report the other side, a capability floor below which small models fail to collaborate, and diminishing returns as agents are added. These protocols hold different quantities constant, and none was built for multi-turn episodes in which a simulated user adds tokens the agent does not control and input dominates the total.

\paragraph{Conversational evaluation and the simulated user.} $\tau$-bench \citep{yao2024tau} scores an agent that converses with a simulated user under a domain policy by the final database state; $\tau^2$-bench \citep{barres2025tau2} adds a telecom domain in which both parties hold tools; in its ablation, moving from a no-user setting, in which the agent controls all tools, to the default dual-control setting lowers GPT-4.1's \passone\ by about 18 points. On $\tau$-bench retail, US human participants succeeded on 45.2\% of a difficulty-balanced task subset, and on the full task set a fixed agent's success moved from 67.0\% to 75.9\% when only the simulator model was swapped \citep{seshadri2026lost}; many general-purpose LLM simulators are over-cooperative and inflate success rates above the human baseline \citep{zhou2026sim2real}. A per-cell MAS effect of the size \citet{kim2026capable} report is within what simulator choice alone can produce, so we fix one simulator and test a second (R2).

\paragraph{Verification of state-changing actions.} On $\tau$-bench traces each deviation in a state-changing action cuts the odds of success by more than 90\% on airline and on retail in the fitted model, while errors in read-only actions matter little \citep{cuadron2025saber}, and dialogue-grounded verifiers add 6 to 12 points of pass$^4$ on $\tau^2$ airline \citep{kang2026policyguard}. These domain-specific designs act on the wrongful-write class that our taxonomy separates from verbal violations. Comparisons of generic architectures on $\tau$-bench-style tasks are scarce and neither compute-matched nor varied over capability \citep{langchain2025bench}; Table~\ref{tab:position} positions this study against the closest work on the properties that matter for testing an architecture-selection rule in a new setting; no prior study has all six at once.

\begin{table}[tbp]
\caption{Position against the closest prior work. No prior study has all six properties at once. ``User'' means a user takes part in the episode, with ``tools'' when the user also holds tools (dual control in the sense of \citet{barres2025tau2}). ``Rule tested'' means an existing selection rule is applied unchanged to a new setting.}
\label{tab:position}
\begin{center}
\footnotesize
\setlength{\tabcolsep}{2pt}
\begin{tabular}{@{}>{\raggedright\arraybackslash}p{2.5cm}>{\raggedright\arraybackslash}p{3.0cm}>{\raggedright\arraybackslash}p{2.3cm}>{\raggedright\arraybackslash}p{3.5cm}>{\raggedright\arraybackslash}p{1.9cm}>{\raggedright\arraybackslash}p{1.3cm}>{\raggedright\arraybackslash}p{1.0cm}@{}}
\toprule
Study & Architectures & Capability varied & Compute matched (what, level) & User (tools) & Sim. & Rule \\
\midrule
\citet{kim2026capable} & SAS + 4 generic MAS & 3 families $\times$ 3 tiers & reasoning tokens, per system, reported as a matched mean; matched maximum iterations & no & n/a & fitted \\
\citet{tran2026single} & SAS + 5 MAS variants & 4 models & thinking tokens, per call, split & no & n/a & no \\
\citet{jwalapuram2026illusion} & CoT-SC vs automated MAS & several & USD, post hoc & no & n/a & no \\
\citet{yu2026bottleneck} & SAS vs MAS & 3 scales & max.\ total steps (experiments); relay bandwidth (theory) & no & n/a & no \\
\citet{gao2025single} & SAS vs MAS & several & cost reported & no & n/a & no \\
\citet{langchain2025bench} & single, swarm, supervisor & 1 model & no & yes (no) & no & no \\
\citet{kang2026policyguard} & agent vs agent + verifier & 3 models & no & yes (no) & no & no \\
This paper & \Sone\ vs \Mone\ (centralized) & 3 families $\times$ 2 tiers & completion tokens, per episode, shared ledger & yes (telecom) & fixed; second on 3 cells & yes \\
\bottomrule
\end{tabular}
\end{center}
\end{table}

%% file: sections/03_design.tex
\section{Experimental design}
\label{sec:design}

\subsection{Setting, systems and unit}
\label{sec:setting}
\label{sec:systems}

\paragraph{Tasks.} We use $\tau^2$-bench v1.0.1 \citep{barres2025tau2} at a pinned commit, with the evaluator and the domain data unmodified; our analysis code never imports the evaluator, and its pass$^k$ reimplementation was checked against the official metric at that commit. The reward checks the final database state and required communication on airline, the database state and natural-language assertions graded by a pinned judge on retail, and the state of the customer's device on telecom (for 7 of its 50 tasks also the actions taken), where the customer runs tools too. Airline uses its 50 base tasks; retail and telecom use fixed 50-task subsets drawn from their base tasks by one seeded shuffle. Only telecom is dual-control in the strict sense of \citet{barres2025tau2}: there the customer holds tools and acts on the device; on airline and retail the agent alone holds tools and the customer acts through the conversation. The original $\tau$-bench was reported to contain tasks that require no state change and lack a check on the required reply, so that an agent that does nothing passes \citep{zhu2025rigorous}, and community corrections followed \citep{cuadron2025saber}; none is applied here, every SAS baseline is generated internally on the pinned version, and no leaderboard number enters a test. Every cell runs every task with three seeds (150 episodes per condition), with at most 200 steps and 10 errors per episode; an episode that ends in an infrastructure error is re-run once and, if it fails again, excluded and reported. One simulator, \texttt{gpt-5.2} at low reasoning effort, plays the customer everywhere; the benchmark's default \texttt{gpt-4.1} is used only in R2. As a reproduction gate before any budget was set, the stock agent with \texttt{gemini-3-flash} on airline under this simulator scored \passone\ 80.7\% against the public leaderboard's 82.5\% for the same configuration (pass$^3$ 74.0\% against 72.0\%), so the pipeline reproduces the published setting.

\paragraph{Models.} Six models from three families, two tiers each: \texttt{gpt-5-nano} and \texttt{gpt-5-mini} (dated August 2025 snapshots, Responses API, reasoning effort high); \texttt{gemini-3.1-flash-lite} and \texttt{gemini-3-flash} (preview, thinking level high); \texttt{qwen3.5-4b}, a 4-billion open-weight model served at 4-bit precision on one laptop with its shipped sampling defaults, a 131k context and one request at a time; and \texttt{qwen3.5-9b} through a hosted router at temperature 0 with reasoning effort high, whose served build is not pinned. The hosted APIs drop the per-trial seed that the benchmark sets; the local route passes it to the model, which matters for the local cells (Appendix~\ref{app:local4b}). The simulator is \texttt{gpt-5.2} (December 2025 snapshot) at low reasoning effort. The run configuration, the prompts and every run's manifest, which records the benchmark commit, the prompt versions and the calibration values in force, are released with the code.

\paragraph{Notation.} A cell $i=(m,d)$ pairs model $m$ with domain $d$ and its task set $\mathcal{T}_d$ ($|\mathcal{T}_d|=50$). For system $A\in\{\Sone,\Mone\}$, task $t$ and seed $s\in\{1,\dots,S\}$ ($S=3$), let $R^{A}_{t,s}\in\{0,1\}$ be the episode reward. Then
\begin{equation}
\label{eq:gain}
P_A(i)=\frac{1}{|\mathcal{T}_d|}\sum_{t\in\mathcal{T}_d}\bar R^{A}_{t},\qquad \bar R^{A}_{t}=\frac{1}{S}\sum_{s=1}^{S}R^{A}_{t,s},\qquad \Delta_i=P_{\Mone}(i)-P_{\Sone}(i).
\end{equation}
The unit is the per-task mean $\bar R^{A}_{t}$, and $P_A(i)$ is \passone, the axis on which the principle is stated: it predicts $\operatorname{sign}\Delta_i=\operatorname{sign}\big(x^{\ast}-P_{\Sone}(i)\big)$ with $x^{\ast}\approx 0.45$. Intervals are task-cluster bootstraps (tasks resampled with their episodes; seed 300), and $\Delta_i$ is reported in points.

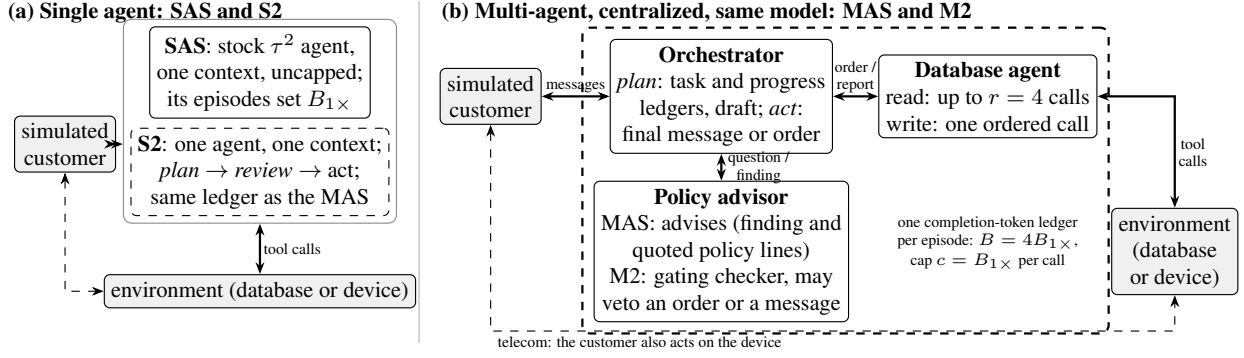
\begin{figure}[t]
\begin{center}
\resizebox{\linewidth}{!}{\input{figures/fig_arch.tex}}
\end{center}
\caption{The systems. (a) Single-agent systems: the stock SAS, whose uncapped episodes calibrate the budget, and S2, one agent that runs the MAS's plan-and-review cycle in one context under the same ledger. (b) The centralized MAS built from the same model: only the orchestrator talks to the customer, only the database agent calls tools, and all calls draw on one completion-token ledger (Eq.~\ref{eq:budget}); in M2 the policy advisor becomes a checker that can veto an order or a message (outlined, not evaluated here). On telecom the customer also acts on the device (dashed).}
\label{fig:design}
\end{figure}

\paragraph{SAS.} The benchmark's stock agent (S1 in the released code): its default system prompt with the domain policy inserted verbatim, native function calling and the highest reasoning effort. This is the SAS over which the principle is stated and the agent behind public leaderboard scores.

\paragraph{MAS.} The centralized architecture adapted to one conversational thread (M1 in the released code; Figure~\ref{fig:design}b), with three agents built from the same model. The \emph{orchestrator} alone talks to the customer and keeps a task ledger and a progress ledger after Magentic-One \citep{fourney2024magentic}; the \emph{database agent} alone calls tools, with up to $r=4$ consecutive calls per read assignment and one call per write, executed as ordered without veto; the \emph{policy advisor} reasons with a fresh context and returns a finding with the quoted policy lines. The benchmark is half-duplex, so each orchestrator turn is a \emph{plan} call without tools and, if the advisor was asked, an \emph{act} call; a turn with a tool call thus costs at least two calls against one in the SAS, the coordination cost the budget charges. Role blocks describe workflow only and follow the unchanged base prompt.

\paragraph{Why centralized.} The other generic architectures of \citet{kim2026capable} need an adaptation that would itself explain the result: in an independent MAS each agent would hold its own conversation with the customer, decentralized debate between customer turns would be confounded by the debate protocol, and hybrid is centralized plus lateral messages. The centralized architecture is the one with a single user-facing locus, which is what the conversation requires, and it is the one whose coordination overhead the budget can charge cleanly.

\paragraph{Control flow.} At the start of an episode one ledger is created with limit $B = 4\Bx$, per-call cap $c = \Bx$, wrap-up threshold $w$ and minimum allowance $a_{\min}$ (\S\ref{sec:budget}). When a customer message or a database-agent report arrives, the orchestrator makes a \emph{plan} call without tools ($\text{max\_tokens} = \min(\text{remaining}, c)$), returning a JSON object with the task-ledger and progress-ledger updates, an optional advisor question and a draft (customer message, read assignment or write assignment). If a question is present the advisor is called with a fresh context and its answer appended, and the orchestrator makes an \emph{act} call that emits the final message or assignment; otherwise the draft is final. A customer message ends the turn. A read assignment hands control to the database agent, which calls tools by native function calling; tool results route to it until it reports or reaches $r = 4$ calls, at which point a report is forced. A write assignment is one call whose executed tool name and arguments are compared with the order and logged (exact, exact after type normalisation, argument mismatch, tool mismatch); a write attempted inside a read assignment is blocked. Every call reserves before dispatch and commits actual usage after; usage is charged before tool-call parsing, so a call cut mid-arguments is charged and its tool calls dropped. A plan that does not parse is retried once, then replaced by a forced customer message; a plan cut at $c$ is not retried. When the remainder falls below $w$ the wrap-up notice is appended once to the system message and only customer messages are allowed; when it falls below $a_{\min}$ the fixed exhaustion message is sent with the label \texttt{budget\_exhausted} and the episode ends (the benchmark records \texttt{agent\_stop}; the analysis reads the label). The simulator's calls are outside the ledger.

\subsection{Budget protocol}
\label{sec:budget}

In the pilot the SAS read 57 input tokens for every token it wrote, so matching total tokens would mostly charge a MAS for duplicated prompts. The comparison is therefore matched on \emph{system-wide completion tokens}: $\kappa_j$ counts the completion tokens of call $j$ by any agent of the system, including hidden reasoning, tool-call arguments and inter-agent messages. Other tokens are reported as deployment accounts, and the meter matches provider billing within 1\% (see the meter checks below).

Budgets are calibrated per cell. The SAS first runs the cell uncapped; these episodes, $\mathcal{E}_{\Sone}(i)$, are the SAS arm and the calibration. For an episode $e$ of the MAS,
\begin{equation}
\label{eq:budget}
C_e=\sum_{j\in e}\kappa_j,\qquad \Bx(i)=Q_{0.75}\big\{C_{e'}:e'\in\mathcal{E}_{\Sone}(i)\big\},\qquad B=k\Bx,
\end{equation}
and call $j$ may write at most $m_j=\min(B-C_{<j},c)$ tokens, where $C_{<j}$ is the spend before it, so all agents draw on one ledger. $k = 4$ is the smallest multiple that left every SAS episode unconstrained in the calibration runs (largest $3.85\Bx$). The per-call cap $c = \Bx$ touches 0 to 1\% of SAS episodes. Once $B-C_{<j}<w=\max(0.15\Bx, 2\,\text{median}_{\Sone})$, a fixed notice tells the system to conclude with the customer; once $B-C_{<j}<a_{\min}$ (the median SAS call), the episode ends with the label \emph{budget exhausted}. Exhausted episodes are scored as they stand and kept apart from inaction; a truncated tool call is charged and never executed. The ledger held in every MAS episode except three on retail $\times$ \texttt{gemini-3.1-flash-lite}, where the provider returned up to 1.8\% more tokens than the limit sent.

Two properties of the protocol matter for reading the results. First, the SAS runs uncapped while the MAS runs under the ceiling, so the asymmetry lies mainly in the notice, and R1 re-runs the SAS capped to size it. Second, the original study matched maximum iteration counts between systems and reports its reasoning-token budgets as a matched mean (about 4,800 tokens per trial), with realised turns of 7.2 for the SAS and 27.7 for the centralized MAS; the text we could read describes no per-episode token ceiling enforced at call time, so its protocol is looser than ours, the grid at $4\Bx$ is a \emph{cost-bounded arm}, and a MAS with no ceiling forms a secondary \emph{uncapped arm} on the three cells at or below 45\%. Table~\ref{tab:protocols} places the protocol against the budget protocols of prior work.

\begin{table}[tbp]
\caption{Relationship to prior budget protocols.}
\label{tab:protocols}
\begin{center}
\footnotesize
\begin{tabular}{@{}>{\raggedright\arraybackslash}p{2.3cm}>{\raggedright\arraybackslash}p{5.2cm}>{\raggedright\arraybackslash}p{7.9cm}@{}}
\toprule
Protocol & Adopted & Deviation here \\
\midrule
\citet{kim2026capable} & per-system ceiling; coordination paid from the same pool; exhaustion is a legal termination & they report matched reasoning-token budgets as a mean per trial and matched maximum iterations, with no per-episode token ceiling described as enforced at call time; here a per-episode token ceiling, enforced \\
\citet{tran2026single} & the SAS spends the whole budget in one trajectory; a MAS divides it & they keep planner and aggregator near budget-neutral, here every component is charged; they count thinking tokens, here completion tokens with reasoning-only as sensitivity; they split statically ($B/k$), here a shared pool with reservations, which is more generous to the MAS, so a null MAS result is conservative \\
\citet{wang2024token} & performance-at-tokens curves & their cap is input plus output per question; here input is reported as a deployment account outside the scientific budget \\
\citet{kapoor2024agents} & accuracy-versus-USD frontier for deployment claims & not used for the main contrast, where compute and architecture would co-vary \\
\bottomrule
\end{tabular}
\end{center}
\end{table}

\paragraph{Implementation.} Each rule above is implemented by one mechanism with a named offline test: one ledger per episode (\texttt{EpisodeBudget} in the agent state; every provider call passes through \texttt{metered\_generate}); reservations bounded by the remainder; per-call limit $\min(\text{reservation}, c)$; wrap-up notice once, on a single system message; no tools after wrap-up; truncated calls charged and never executed; the exhaustion label distinct from inaction; provider overrun recorded; utilisation, wrap-up, exhaustion and truncation rates computable from every episode's snapshot; base-prompt injection with a check that the base prompt is unchanged; MAS routing as above. The suite holds 65 tests at the pinned commit on the benchmark's mock domain, including one full MAS episode, one capped SAS episode and one S2 episode through the benchmark's own runner. The uncapped arm sets no limit and no wrap-up, keeps $c$, and removes the orchestrator's allowance paragraph; a test checks that the prompt difference is exactly that paragraph. A flag added after the grid was read lets the orchestrator see the tool definitions with tool choice disabled (\S\ref{sec:compliance}); it is off by default and off in every cell reported here.

\paragraph{Meter checks.} The meter was checked on the reproduction gate run before any budget was set. Reasoning-token visibility: on all 1,614 Gemini agent calls carrying a usage-details block, $\text{reasoning} + \text{text} = \text{completion}$ exactly (698,493 reasoning and 144,363 text tokens; reasoning is 82\% of output); \texttt{gpt-5.2} likewise (10,466 reasoning inside 59,538 completion). 123 calls (7\%) carry no details block, covering 1.2\% of output tokens, which the primary meter counts and a reasoning-only meter would miss. Cached-input accounting: $\text{cached} + \text{text} = \text{prompt}$ per call; recomputing cost at three rates reproduces the recorded totals to the cent for both providers, whereas flat input pricing would overstate the Gemini agent cost by 42\%. Billing: recorded \$6.87 against \$6.86 billed (Google, $-0.1$\%) and \$1.67 against \$1.69 (OpenAI, $+0.9$\%). Regrade determinism: re-evaluating 300 baseline episodes with the assertion judge forced on reproduced every original reward. Task-set identity: byte-identical task definitions and assertion lists across runs. Input-to-output ratio: about 60:1 for the non-reasoning pilot model and 16.9:1 for the reasoning gate run, so the completion-token meter diverges from total tokens far more for non-reasoning models; both are reported. On the local route the completion count includes thinking tokens and the thinking trace never reaches the customer; on the hosted router of the 9-billion model the per-call limit applied to visible text only (\S\ref{sec:compliance}).

\paragraph{Calibration.} Table~\ref{tab:calibration} gives, for every grid cell and for the October-build retail \texttt{qwen3.5-4b} cell (Appendix~\ref{app:local4b}), the SAS spend that set the budget and the budget values the MAS ran under.

\begin{table}[tbp]
\caption{Calibration per cell from the 150 uncapped \Sone\ episodes: mean and 75th percentile of episode completion tokens ($\Bx$), the ceiling $B = 4\Bx$, the per-call cap as run ($c = \Bx$) and the \Sone\ per-call 95th percentile for reference. The wrap-up threshold is $w = \max(0.15\Bx, 2\,\text{median})$ and the minimum allowance $a_{\min} = \text{median}$ of \Sone\ per-call completion tokens (full values in the released \texttt{budgets.json}).}
\label{tab:calibration}
\begin{center}
\footnotesize
\setlength{\tabcolsep}{5pt}
\begin{tabular}{@{}llrrrrr@{}}
\toprule
Domain & Model & \Sone\ mean & $\Bx$ ($P_{75}$) & $B = 4\Bx$ & $c$ & per-call $P_{95}$ \\
\midrule
airline & \texttt{gpt-5-nano} & 40,839 & 51,297 & 205,188 & 51,297 & 11,566 \\
retail & \texttt{gpt-5-nano} & 29,783 & 35,313 & 141,252 & 35,313 & 8,897 \\
telecom & \texttt{gpt-5-nano} & 60,570 & 75,032 & 300,128 & 75,032 & 6,100 \\
airline & \texttt{gpt-5-mini} & 20,808 & 27,477 & 109,908 & 27,477 & 8,312 \\
retail & \texttt{gpt-5-mini} & 14,432 & 18,294 & 73,176 & 18,294 & 5,389 \\
telecom & \texttt{gpt-5-mini} & 26,165 & 31,514 & 126,056 & 31,514 & 5,273 \\
airline & \texttt{gemini-3.1-flash-lite} & 4,997 & 6,159 & 24,636 & 6,159 & 1,662 \\
retail & \texttt{gemini-3.1-flash-lite} & 3,464 & 4,186 & 16,744 & 4,186 & 949 \\
telecom & \texttt{gemini-3.1-flash-lite} & 5,272 & 7,123 & 28,492 & 7,123 & 1,326 \\
airline & \texttt{gemini-3-flash} & 5,687 & 6,672 & 26,688 & 6,672 & 1,813 \\
retail & \texttt{gemini-3-flash} & 3,471 & 4,161 & 16,644 & 4,161 & 1,072 \\
telecom & \texttt{gemini-3-flash} & 7,242 & 8,954 & 35,816 & 8,954 & 1,355 \\
telecom & \texttt{qwen3.5-4b} & 4,335 & 5,630 & 22,520 & 5,630 & 608 \\
retail & \texttt{qwen3.5-4b} (Oct.\ build) & 4,461 & 5,369 & 21,476 & 5,369 & 1,179 \\
airline & \texttt{qwen3.5-9b} & 9,624 & 12,263 & 49,052 & 12,263 & 3,243 \\
retail & \texttt{qwen3.5-9b} & 5,367 & 6,370 & 25,480 & 6,370 & 1,808 \\
telecom & \texttt{qwen3.5-9b} & 4,514 & 5,758 & 23,032 & 5,758 & 772 \\
\bottomrule
\end{tabular}
\end{center}
\end{table}

\subsection{Outcome measures}
\label{sec:measures}

We use three measures, each blind to something the others catch. The \emph{primary} measure is the official reward, which on airline and telecom misses violations that never reach a write, such as promising a forbidden action. The \emph{secondary} measure is episode-level compliance with every natural-language assertion under the official judge. The \emph{tertiary} measure is the pass rate on tasks whose gold trajectory contains a write, because the database comparison rewards passivity on refusal tasks.

\paragraph{Failure taxonomy.} Per episode and condition: \emph{WW}, wrongful write, a write the gold trajectory does not contain (visible to the primary measure); \emph{IN}, inaction, the gold trajectory requires a write and the agent made none; \emph{BX}, budget exhausted, the episode ended on the ledger's exhaustion path (kept apart from IN; the inaction rate is reported with and without it); \emph{FP}, a passing episode that the assertion judge flags for a policy violation in words (defined for airline, where the assertions are outside the official score; telecom tasks carry none and on retail they are inside the score); \emph{order misexecuted} (MAS episodes only), a write whose executed tool name or arguments differ from the orchestrator's assignment, computed from logs; \emph{other}. Write tools are read from the benchmark's own \texttt{WRITE} tags; only agent-side calls count (telecom customers call tools too); gold actions are filtered to the assistant requestor. Failing episodes are coded in the order WW, BX, IN, other.

\paragraph{The judge as an instrument.} The judge (\texttt{gpt-4.1}, April 2025 snapshot, temperature 0, the benchmark's pinned default) was validated before use: a full-population regrade flagged exactly the same 12 of 121 passing gate-run episodes as an earlier passing-only screen; assertion violations are about 6.5 times more frequent in failing than in passing gate-run episodes (65.5\% against 9.9\%), so the ungraded dimension is concordant with the graded one. Commission-type flags (the assertion text begins ``Agent does not / should not / must not / should never / never / should refuse'') are taken as-is, while omission-type flags (duties to detect, verify, identify or communicate) were not adjudicated by hand and enter only an upper-bound count; the judge was not checked against human labels. The measured blind spot of the primary measure on the two anchor runs is 15.0\% of passing episodes at a 40\% baseline and 9.9\% at 80.7\% (upper bounds). Anchor values for H2a from the two stock-prompt baseline runs: WW/failures 41\% at a 40.0\% baseline and 7\% at 80.7\%; FP/passes 15\% and 10\%.

\subsection{Hypotheses, validity checks and power}
\label{sec:hypotheses}

Two hypotheses are primary; everything else is secondary. Table~\ref{tab:card} (Appendix~\ref{app:card}) gives each test, its reading rule and the result in one row.

\textbf{H1.} The primary model regresses the gain on a capability index,
\begin{equation}
\label{eq:h1}
\Delta_i=\alpha+\beta x_i+\gamma_{f(i)}+\delta_{d(i)}+\varepsilon_i,\qquad x_i=\frac{1}{2}\sum_{d'\neq d}P_{\Sone}(m,d'),
\end{equation}
with family and domain fixed effects $\gamma_{f(i)}$ and $\delta_{d(i)}$, fitted by weighted least squares with weights $\omega_i=\widehat{\operatorname{Var}}(\Delta_i)^{-1}$ from the task-cluster bootstrap. The index $x_i$, the model's SAS \passone\ on its other two domains, shares no episode with $\Delta_i$; Lemma~\ref{lem:coupling} shows why this matters. H1 holds if $\hat\beta<0$ with a 95\% CI excluding zero, with the $p$-value from 10,000 permutations of $\Delta_i$ within family. The crossing at the mean fixed effect, $x_0=-(\alpha+\bar\gamma+\bar\delta)/\beta$, with a task-cluster bootstrap CI, gives the reading: the principle \emph{transfers} if H1 holds and the CI of $x_0$ covers $0.45$, \emph{shifts} if H1 holds and it does not, and \emph{disappears} if H1 fails. Split-seed, logit and relative-gain forms are reported beside it, and the naive form $x_i=P_{\Sone}(i)$ descriptively.

\begin{lemma}[Coupling]
\label{lem:coupling}
Let $P_{\Sone}(i)=p_i+e_i$ and $\Delta_i=g+\beta p_i+u_i-e_i$, with latent capability $p_i$ and independent mean-zero sampling errors $e_i$, $u_i$ (variances $\sigma_p^2$, $\sigma_e^2$, $\sigma_u^2$). The least-squares slope of $\Delta_i$ on $P_{\Sone}(i)$ converges to $\lambda\beta-(1-\lambda)$, where $\lambda=\sigma_p^2/(\sigma_p^2+\sigma_e^2)$; on an index $x_i=p_i+e'_i$ built from other episodes, with error variance $\sigma_{e'}^2$, it converges to $\lambda'\beta$, where $\lambda'=\sigma_p^2/(\sigma_p^2+\sigma_{e'}^2)$.
\end{lemma}
\begin{proof}
Write $y_i=\Delta_i=g+\beta p_i+u_i-e_i$. For a single regressor $z_i$, the least-squares slope converges to $\operatorname{Cov}(y,z)/\operatorname{Var}(z)$. For the naive regressor $z_i=P_{\Sone}(i)=p_i+e_i$, independence gives $\operatorname{Cov}(y,z)=\beta\sigma_p^2-\sigma_e^2$ and $\operatorname{Var}(z)=\sigma_p^2+\sigma_e^2$, so the slope converges to
\begin{equation*}
\frac{\beta\sigma_p^2-\sigma_e^2}{\sigma_p^2+\sigma_e^2}=\lambda\beta-(1-\lambda).
\end{equation*}
For $z_i=x_i=p_i+e'_i$, the error $e'_i$ is independent of $(p_i,e_i,u_i)$, so $\operatorname{Cov}(y,z)=\beta\sigma_p^2$ and $\operatorname{Var}(z)=\sigma_p^2+\sigma_{e'}^2$, and the slope converges to $\lambda'\beta$. With fixed effects, the Frisch--Waugh--Lovell theorem reduces the fit to a single-regressor fit on residuals after partialling out the family and domain indicators, and the argument applies to the residualised variables, with $\sigma_p^2$, $\sigma_e^2$ and $\sigma_{e'}^2$ read as within-group variances; weights rescale these variances without changing the signs. If the two systems' errors are correlated, as when both run the same tasks, with $\operatorname{Cov}(u_i,e_i)=\sigma_{ue}$, the naive limit becomes $\lambda\beta-(\sigma_e^2-\sigma_{ue})/(\sigma_p^2+\sigma_e^2)$, still shifted down whenever $\sigma_{ue}<\sigma_e^2$, while the out-of-domain limit is unchanged, because $e'_i$ comes from episodes on other domains.
\end{proof}
The naive slope is therefore negative even when $\beta=0$, the coupling artefact of change-score regression \citep{tu2007change}, while an out-of-domain index can hide a slope but never create one. The sign match, the bridge to the original 15 of 16, counts cells whose SAS CI does not straddle 45\% and is secondary, since a sign is a discontinuous metric that can manufacture thresholds \citep{schaeffer2023mirage}.

\textbf{H2.} Stated as rates, because the share $\text{FP}/(\text{FP}+\text{WW})$ rises mechanically with the pass rate. H2a, on the SAS: $\text{WW}/\text{failures}$ declines with the index while $\text{FP}/\text{passes}$ does not. H2b, within cells: the paired change in WW, always read with the changes in IN, BX and FP, so that a drop in WW bought by passivity or exhaustion is read as one failure traded for another.

\textbf{Validity (G1).} A null H1 may mean the MAS is broken \citep{cemri2025mast}, so every MAS cell is checked from logs, never from reward. Four mechanical bars: calls cut at the cap ($\le 5\%$), tokens in cut calls ($\le 10\%$), no turn with two cut calls outside the final exhaustion turn (the exemption of the exhaustion turn is an amendment adopted after the cells were read; \S\ref{sec:compliance}), and no JSON or cut text sent to the customer. One fidelity bar: at least 90\% of the orchestrator's write orders executed exactly as issued. Exhaustion is an outcome, never a validity failure; only bug-class fixes are allowed, a protocol failure calls for a fix and a re-run, and a family that still fails leaves H1. How the fidelity bar fared is reported in \S\ref{sec:compliance}.

\textbf{Robustness.} R1 re-runs one cell below and one above 45\% at other budgets with the SAS capped; R2 re-runs three airline cells, the straddling cell and two above 45\%, with the benchmark's default simulator. A strong SAS (S2) that runs the MAS's plan-and-review cycle in one context under the same ceiling is run on three cells as an extension (\S\ref{sec:extension}).

\textbf{Power.} A simulation (\texttt{power\_sim.py}; 400 worlds per setting, seed 1, 2 September 2026) drew cell gains on a line with task heterogeneity fitted to the pilot and gate runs (beta-binomial: pilot mean 0.400, $\kappa = 1.39$; gate mean 0.807, $\kappa = 0.42$; $\kappa = 0.91$ used). Power is the fraction of worlds in which the slope CI excludes zero on the negative side; the $b = 0$ column of Table~\ref{tab:power} is the false-positive rate. The leave-one-domain-out form gives power 0.94 at $\beta=-0.30$ and 0.74 at $-0.20$ for an 18-cell grid, so a null is read as $|\beta|<0.20$. A replica of the original 16-cell validation (20 tasks, one run) has expected sign accuracy 0.74 under $b = -0.30$ and 0.60 under the null, and reaches 15 of 16 in 4\% of true-effect worlds (1\% under the null). H2a (WW/failures on the index): power 1.00 for the anchor effect, false-positive rate 0.03. The simulation modelled task-level noise only; what it did not model is reported with the result (\S\ref{sec:h1}).

\begin{table}[tbp]
\caption{Power of the H1 slope test by form of $x$ and grid size (simulation). acc.: expected sign accuracy at $b = -0.3$; inform.: share of informative cells.}
\label{tab:power}
\begin{center}
\footnotesize
\setlength{\tabcolsep}{4pt}
\begin{tabular}{@{}llrrrrrrr@{}}
\toprule
$x$ & Scenario & $b = 0$ & $-0.10$ & $-0.20$ & $-0.30$ & $-0.40$ & acc.\ @$-0.3$ & inform.\ @$-0.3$ \\
\midrule
naive & 27 cells, 50 tasks, 3 seeds & 0.19 & 0.78 & 1.00 & 1.00 & 1.00 & 0.84 & 0.39 \\
naive & 18 cells, 50 tasks, 3 seeds & 0.20 & 0.67 & 0.99 & 1.00 & 1.00 & 0.87 & 0.49 \\
split seed & 27 cells, 50 tasks, 3 seeds & 0.07 & 0.48 & 0.87 & 0.99 & 1.00 & 0.82 & 0.32 \\
split seed & 18 cells, 50 tasks, 3 seeds & 0.07 & 0.35 & 0.80 & 0.97 & 1.00 & 0.86 & 0.42 \\
leave-one-domain-out & 27 cells, 50 tasks, 3 seeds & 0.07 & 0.40 & 0.77 & 0.95 & 1.00 & 0.84 & 0.38 \\
leave-one-domain-out & 27 cells, 50 tasks, 5 seeds & 0.10 & 0.51 & 0.93 & 0.99 & 1.00 & 0.86 & 0.47 \\
leave-one-domain-out & 18 cells, 50 tasks, 3 seeds & 0.07 & 0.37 & 0.74 & 0.94 & 0.98 & 0.88 & 0.50 \\
\bottomrule
\end{tabular}
\end{center}
\end{table}

%% file: figures/fig_arch.tex
\begin{tikzpicture}[
  font=\scriptsize,
  agent/.style={draw, rounded corners=2pt, align=center, inner sep=2.5pt, fill=white},
  ext/.style={draw, rounded corners=2pt, align=center, inner sep=2.5pt, fill=black!6},
  arr/.style={{Stealth[length=1.5mm]}-{Stealth[length=1.5mm]}, thick},
  lab/.style={font=\fontsize{5.6}{6.2}\selectfont, align=center, inner sep=1pt}
]
\node[anchor=west, font=\scriptsize\bfseries] at (-0.85,1.05) {(a) Single agent: SAS and S2};
\node[ext] (c1) at (0,-0.6) {simulated\\customer};
\node[agent] (sas) at (2.45,0.3) {\textbf{SAS}: stock $\tau^2$ agent,\\one context, uncapped;\\its episodes set $\Bx$};
\node[agent, dashed] (s2) at (2.45,-0.95) {\textbf{S2}: one agent, one context;\\\emph{plan} $\rightarrow$ \emph{review} $\rightarrow$ act;\\same ledger as the MAS};
\node[draw, black!45, rounded corners=3pt, fit=(sas)(s2), inner sep=2.5pt] (single) {};
\node[ext] (e1) at (2.45,-2.45) {environment (database or device)};
\draw[arr] (c1) -- (c1 -| single.west);
\draw[arr] (single.south) -- node[lab, right] {tool calls} (e1.north);
\draw[arr, dashed, thin] (c1.south) |- (e1.west);

\draw[black!30] (4.45,1.2) -- (4.45,-3.1);

\node[anchor=west, font=\scriptsize\bfseries] at (4.6,1.05) {(b) Multi-agent, centralized, same model: MAS and M2};
\node[ext] (c2) at (5.35,0) {simulated\\customer};
\node[agent] (orch) at (8.25,0) {\textbf{Orchestrator}\\\emph{plan}: task and progress\\ledgers, draft; \emph{act}:\\final message or order};
\node[agent] (adv) at (8.25,-1.95) {\textbf{Policy advisor}\\MAS: advises (finding and\\quoted policy lines)\\M2: gating checker, may\\veto an order or a message};
\node[agent] (db) at (11.6,0) {\textbf{Database agent}\\read: up to $r=4$ calls\\write: one ordered call};
\node[ext] (e2) at (13.95,-1.95) {environment\\(database\\or device)};
\draw[arr] (c2) -- node[lab, above] {messages} (orch);
\draw[arr] (orch) -- node[lab, right, xshift=1pt] {question /\\finding} (adv);
\draw[arr] (orch) -- node[lab, above] {order /\\report} (db);
\draw[arr] (db.east) -| node[lab, pos=0.75, right] {tool\\calls} (e2.north);
\node[draw, thick, dashed, rounded corners=3pt, fit=(orch)(adv)(db), inner sep=4pt] (ledger) {};
\node[lab, align=center] at (11.6,-1.85) {one completion-token ledger\\per episode: $B=4\Bx$,\\cap $c=\Bx$ per call};
\draw[arr, dashed, thin] (c2.south) -- (c2.south |- 0,-2.95) -| (e2.south);
\node[lab, anchor=north west] at (5.4,-2.97) {telecom: the customer also acts on the device};
\end{tikzpicture}

%% file: sections/04_results.tex
\section{Results}
\label{sec:results}

\subsection{The grid as run}
\label{sec:compliance}

\paragraph{Scope.} Of the 18 cells, 16 were run on the registered configuration and form the grid of Table~\ref{tab:cells}. The registered build of \texttt{qwen3.5-4b} became unavailable before its retail cell ran; that cell was run on a later build of the same model, with its own SAS calibration, and is reported outside every fit (Appendix~\ref{app:local4b}). The airline \texttt{qwen3.5-4b} cell was not run by us; a collaborator's MAS run of it enters only as a sensitivity (Appendix~\ref{app:local4b}).

\paragraph{Cells below 45\%.} Two cells are clearly below 45\%, one straddles and thirteen are clearly above (Table~\ref{tab:cells}). The weakest open model, \texttt{qwen3.5-4b} served at 4-bit, scores 94.7\% on telecom, above every closed model (\texttt{qwen3.5-9b} scores 95.3\%), and adds no low cell. Telecom separates the models most: the two low-tier closed models sit at 32.7 and 34.0\%, \texttt{gpt-5-mini} at 77.3\%, and \texttt{gemini-3-flash} and both open models at 91.3 to 95.3\%. The principle's distinctive prediction, a gain below 45\%, is therefore tested on two cells.

\paragraph{Validity and one deviation from the plan.} The twelve OpenAI and Google MAS cells and telecom $\times$ \texttt{qwen3.5-4b} pass the four mechanical bars: calls cut, tokens in cut calls, turns with two cut calls outside the exhaustion turn, and JSON or cut text sent to the customer (Table~\ref{tab:validity}). The fidelity bar was not met in any cell. Under the stricter of our two audits, which counts each unique order once and counts a refused order as not executed, write orders were executed exactly as issued in 9 to 41\% of cases per cell (680 of 2,744, 25\%, over the 13 cells): the orchestrator plans without the tool schemas, which only the database agent holds, so 13\% of its orders named a tool that does not exist and were mapped to the right one, 19\% carried argument names the tool does not take and were corrected, and 41\% were refused or diverted and reported back (Table~\ref{tab:validity}). The pre-registered response to a protocol failure is a bug-class fix and a re-run. The fix was implemented after the grid had been read: the orchestrator receives the tool definitions with tool choice disabled, so it still cannot call a tool. On a smoke of 5 tasks on each of three cells it raised the share of orders executed as issued from about 25\% to 72\% (13 of 18), short of the bar, with two of the five misses being refusals by the database agent on policy grounds, which no orchestrator-side change touches. The grid was not re-run. We therefore report the MAS as built and label this a deviation from the plan; throughout, ``the 13 cells'' means the 13 cells that meet the mechanical bars, and we read the unexecuted orders as part of what dividing the work costs in this architecture, which limits what the results say about designs that share tool schemas (\S\ref{sec:discussion}). \texttt{qwen3.5-4b} failed to parse 56\% of its plan calls; excluding its cell leaves the pattern of every result unchanged (sign match 9 of 11, binomial $p = 0.065$; pooled WW change $-0.7$ points, CI $[-2.9, +1.3]$). The three \texttt{qwen3.5-9b} cells fail the mechanical bars, all on turns with two cut calls and two on tokens in cut calls; a larger cap for one family would change the protocol, so they enter only a sensitivity. The \texttt{qwen3.5-4b} cell has no leave-one-domain-out index, so the primary fit uses 12 cells.

\begin{table}[t]
\caption{\textbf{No cell shows a MAS benefit.} The grid in the cost-bounded arm ($B = 4\Bx$, $c = \Bx$, $r = 4$; 150 episodes per system), \passone\ in percent with task-cluster bootstrap CIs. Exh.: MAS episodes ending budget exhausted; util.: median share of the ceiling spent; fin./exh.: pass rates of MAS episodes that finished and that were exhausted; tokens: median MAS completion tokens per episode over the SAS median (the ceiling bounds this ratio at $B$ over the SAS median, 5.0 to 6.3 across cells). $\dagger$: fails the mechanical bars, sensitivity only; SAS values computed by a collaborator. $\ddagger$: local route; per trial the MAS passed 15, 11 and 9 of 50, with the third trial affected by the route's fixed sampling seed (Appendix~\ref{app:local4b}).}
\label{tab:cells}
\begin{center}
\scriptsize
\setlength{\tabcolsep}{2.5pt}
\begin{tabular}{@{}llrlrlrrcr@{}}
\toprule
Model & Domain & SAS & SAS 95\% CI & MAS & Gain [95\% CI] & Exh. & Util. & Fin./exh. & Tokens \\
\midrule
\texttt{gpt-5-nano} & airline & 42.7 & [31.3, 54.0]$^{\circ}$ & 45.3 & $+2.7$ [$-6.7$, $+12.7$] & 23\% & 0.50 & 50\% / 31\% & 2.9$\times$ \\
 & retail & 68.0 & [58.7, 77.3] & 44.7 & $-23.3$ [$-32.7$, $-13.3$] & 16\% & 0.63 & 52\% / 8\% & 3.1$\times$ \\
 & telecom & 32.7 & [22.7, 43.3]$^{\downarrow}$ & 19.3 & $-13.3$ [$-20.7$, $-6.0$] & 15\% & 0.54 & 20\% / 13\% & 2.8$\times$ \\
\addlinespace[2pt]
\texttt{gpt-5-mini} & airline & 74.7 & [64.7, 84.0] & 58.0 & $-16.7$ [$-27.3$, $-6.7$] & 34\% & 0.83 & 68\% / 39\% & 5.2$\times$ \\
 & retail & 76.0 & [66.7, 84.0] & 48.0 & $-28.0$ [$-40.0$, $-17.3$] & 43\% & 0.90 & 68\% / 22\% & 4.9$\times$ \\
 & telecom & 77.3 & [68.0, 86.0] & 66.0 & $-11.3$ [$-18.7$, $-4.0$] & 18\% & 0.70 & 79\% / 7\% & 4.2$\times$ \\
\addlinespace[2pt]
\texttt{gemini-3.1-flash-lite} & airline & 78.7 & [68.0, 88.7] & 62.7 & $-16.0$ [$-24.0$, $-8.0$] & 31\% & 0.71 & 85\% / 13\% & 4.1$\times$ \\
 & retail & 77.3 & [67.3, 86.7] & 41.3 & $-36.0$ [$-47.3$, $-24.0$] & 58\% & 1.00 & 75\% / 17\% & 5.2$\times$ \\
 & telecom & 34.0 & [24.0, 44.7]$^{\downarrow}$ & 12.0 & $-22.0$ [$-30.7$, $-12.7$] & 45\% & 0.96 & 20\% / 1\% & 5.8$\times$ \\
\addlinespace[2pt]
\texttt{gemini-3-flash} & airline & 80.7 & [70.7, 90.0] & 55.3 & $-25.3$ [$-36.7$, $-14.0$] & 51\% & 1.00 & 74\% / 37\% & 6.1$\times$ \\
 & retail & 80.0 & [70.7, 88.7] & 25.3 & $-54.7$ [$-66.0$, $-44.0$] & 67\% & 1.00 & 63\% / 7\% & 5.1$\times$ \\
 & telecom & 91.3 & [86.0, 96.0] & 10.7 & $-80.7$ [$-87.3$, $-73.3$] & 68\% & 1.00 & 33\% / 0\% & 5.4$\times$ \\
\addlinespace[2pt]
\texttt{qwen3.5-4b}$^{\ddagger}$ & telecom & 94.7 & [90.7, 98.0] & 23.3 & $-71.3$ [$-80.0$, $-62.7$] & 63\% & 1.00 & 54\% / 5\% & 5.3$\times$ \\
\addlinespace[2pt]
\texttt{qwen3.5-9b}$^{\dagger}$ & airline & 79.3 & [71.3, 86.7] & 61.3 & $-18.0$ [$-28.7$, $-6.7$] & 12\% & -- & -- & -- \\
 & retail & 76.0 & [66.7, 84.7] & 49.3 & $-26.7$ (no paired CI) & 10\% & 0.51 & 55\% / 0\% & -- \\
 & telecom & 95.3 & [91.3, 98.7] & 33.3 & $-62.0$ (no paired CI) & 33\% & 0.93 & 48\% / 4\% & -- \\
\bottomrule
\multicolumn{10}{@{}l}{$^{\downarrow}$ SAS CI entirely below 45\%; $^{\circ}$ SAS CI straddles 45\%; every other cell is entirely above. Airline and retail \texttt{qwen3.5-4b}: Appendix~\ref{app:local4b}.}
\end{tabular}
\end{center}
\end{table}

\begin{table}[tbp]
\caption{G1 indicators per MAS cell at $4\times$, computed from logs. Bars: calls cut at the cap $\le 5\%$; tokens in cut calls $\le 10\%$; no turn with two or more cut calls outside the final exhaustion turn (shown as exhaustion / other); no JSON or cut text sent to the customer. Reported indicators: episodes (of 150) that consulted the advisor; plan calls that did not parse (share and count); forced customer messages; and the share of unique write orders executed exactly as issued under the stricter audit (a refused or diverted order counts as not executed). The fidelity bar ($\ge 90\%$ as issued) fails in every cell. Last column: whether the mechanical bars are met. ``--'': not available from the collaborator's files.}
\label{tab:validity}
\begin{center}
\scriptsize
\setlength{\tabcolsep}{3pt}
\resizebox{\linewidth}{!}{%
\begin{tabular}{@{}llccccrrrcl@{}}
\toprule
Model & Domain & Cut calls & Cut tokens & 2-cut (exh./other) & JSON/cut text & Advisor & Plan fail (n) & Forced & As issued & Bars \\
\midrule
\texttt{gpt-5-nano} & airline & 0.3\% & 0.3\% & 0 / 0 & 0 & 17 & 3.5\% (48) & 31 & 16\% & met \\
 & retail & 0.4\% & 0.5\% & 0 / 0 & 0 & 13 & 1.1\% (15) & 43 & 10\% & met \\
 & telecom & 0.3\% & 0.5\% & 0 / 0 & 0 & 9 & 1.6\% (30) & 51 & 20\% & met \\
\texttt{gpt-5-mini} & airline & 0.9\% & 0.9\% & 0 / 0 & 0 & 21 & 3.9\% (46) & 40 & 31\% & met \\
 & retail & 1.3\% & 1.4\% & 0 / 0 & 0 & 7 & 4.8\% (58) & 54 & 39\% & met \\
 & telecom & 0.5\% & 0.5\% & 0 / 0 & 0 & 8 & 1.9\% (32) & 25 & 41\% & met \\
\texttt{gemini-3.1-flash-lite} & airline & 1.8\% & 2.5\% & 0 / 0 & 0 & 93 & 3.7\% (37) & 38 & 13\% & met \\
 & retail & 3.5\% & 6.7\% & 2 / 0 & 0 & 106 & 6.6\% (68) & 78 & 13\% & met \\
 & telecom & 1.8\% & 2.0\% & 0 / 0 & 0 & 130 & 4.0\% (39) & 39 & 15\% & met \\
\texttt{gemini-3-flash} & airline & 1.9\% & 3.0\% & 1 / 0 & 0 & 147 & 4.9\% (48) & 46 & 41\% & met \\
 & retail & 3.3\% & 5.8\% & 1 / 0 & 0 & 150 & 5.2\% (46) & 42 & 19\% & met \\
 & telecom & 1.7\% & 1.5\% & 0 / 0 & 0$^{a}$ & 150 & 6.0\% (66) & 64 & 29\% & met \\
\texttt{qwen3.5-4b} & telecom & 2.3\% & 8.0\% & 0 / 0 & 0$^{a}$ & 14 & 56.2\% (1,386) & 1,076 & 9\% & met \\
\texttt{qwen3.5-4b}$^{c}$ & retail (Oct.\ build) & 3.6\% & 15.6\% & 7 / 5 & 0 & 13 & 21.3\% (347) & 80 & 16\% & fails \\
\texttt{qwen3.5-4b}$^{d}$ & airline (collab.) & 3.2\% & 12.9\% & 3 / 1 & 0 & 22 & 21.1\% (360) & 90 & 33\% & fails \\
\texttt{qwen3.5-9b} & airline$^{b}$ & -- & 35.7\% & 5 / 9 & 0 & -- & 33\% (492) & -- & -- & fails \\
 & retail & -- & 9.6\%$^{e}$ & 1 / 4 & 0 & 59 & 25\% (352) & 144 & -- & fails \\
 & telecom & 3.2\%$^{f}$ & 12.9\% & 5 / 6 & 0 & 52 & 14\% (292) & 122 & -- & fails \\
\bottomrule
\end{tabular}}
\end{center}
\vspace{2pt}
{\footnotesize\setlength{\parindent}{0pt} Notes. $^{a}$ One \texttt{cut text} flag raised by the checker, inspected and found to be a false positive: a plan call with \texttt{finish\_reason: length} and a complete JSON object; the customer message was complete. $^{b}$ Run by a collaborator on the same configuration at a later commit of the benchmark, outside the pinned commit's history. $^{c}$ Run on the October build of the model, outside every fit (Appendix~\ref{app:local4b}). $^{d}$ Run by a collaborator on the registered build against a budget calibrated on a SAS from a different serving stack; enters only as a sensitivity (Appendix~\ref{app:local4b}). Double cuts: the three non-zero entries among the 13 cells that meet the bars are all on exhaustion turns, which the amended bar allows (amendment adopted 17 October 2026, after the cells were read; no cell changes status under it). Write orders, stricter audit: 2,744 unique orders over the 13 cells; 24.8\% executed as issued; 13.1\% named a tool that does not exist and were mapped to the right one; 19.4\% carried argument names the tool does not take and were corrected; 41.1\% refused or diverted and reported back; 1.5\% executed with a changed value. $^{e}$ 13.2\% excluding runaway calls. $^{f}$ Share of plan calls. The looser audit of the plan (2,771 closed assignments, an order counted by what the database agent then executed) gives 28\% as issued, 36\% executed with a different tool or arguments, 35\% not executed, 0.6\% not verifiable. Fix smoke (orchestrator shown the tool definitions, tool choice disabled; 5 tasks on each of three cells, one trial): 13 of 18 orders as issued (1 of 4, 7 of 9, 5 of 5); the misses were two policy refusals by the database agent, one parallel-call wrapper, one customer-side tool and one corrected key.\par}
\end{table}

\paragraph{Read-call limit.} The database agent's limit of $r = 4$ was reached on 35 / 26 / 17\% of read orders (airline / retail / telecom) for \texttt{gemini-3.1-flash-lite}, 50 / 36 / 51\% for \texttt{gemini-3-flash}, 68 / 55 / 44\% for \texttt{gpt-5-mini} and 30 / 72\% for \texttt{gpt-5-nano} on retail / telecom, so it bound on 17 to 72\% of read orders in the 11 cells where it was counted. Re-running two telecom cells at $r = 25$ left both gains inside their $r = 4$ intervals ($-21.3$ against $-22.0$; $-77.3$ against $-80.7$) and plan calls per episode flat, so the overhead sits in the orchestrator's planning.

\paragraph{The hosted route of the 9-billion model.} The route returned more than \texttt{max\_tokens} on 17 (telecom) and 12 (retail) calls, of which 4 and 6 were runaway calls of about 131k reasoning tokens that exhausted the episode in one call, while the collaborator's retail replication had 6 such calls and no runaway, so the route's handling of the reasoning budget varies by run; \passone\ excluding those episodes is 34.2\% [27.0, 42.3] on telecom and 51.4\% [43.3, 59.4] on retail. Retail $\times$ \texttt{qwen3.5-9b} run twice with the same agent settings and budgets (ours at concurrency 4 on the pinned commit; the collaborator's at concurrency 1 on a later benchmark commit) gave 49.3\% and 51.3\%.

\subsection{The grid: no cell shows a MAS benefit}
\label{sec:grid}

Table~\ref{tab:cells} gives the grid before any regression. The MAS is below the SAS in 12 of the 13 cells, with a gain CI excluding zero in each, and ties in the straddling cell; the three cells that fail the mechanical bars show the same sign. Within families the loss grows with the strength of the SAS, and the largest losses fall on the two cells where the SAS exceeds 90\%. Scored against the SAS on the same task, exhausted MAS episodes carry 74\% of the mean loss over the 13 cells ($-22.4$ of $-30.5$ points, an exact split; Table~\ref{tab:loss}). In ten cells the loss runs mainly through the budget, 60 to 90\% of it via exhaustion: 18 to 68\% of MAS episodes end exhausted and pass at 0 to 39\%. On the three \texttt{gpt-5-nano} cells it runs through decisions: on retail and telecom, finished episodes carry 73\% of the loss, and on telecom the MAS hands 120 customers to a human agent against 79. Per task, the MAS is better on 43 of 650 task-cell pairs and worse on 330, and no task improves for more than two of the four closed models.

\paragraph{Where the loss arises.} For each MAS episode on task $t$ we take its reward minus $\bar R^{\Sone}_{t}$, the SAS pass rate on the same task, and average over the MAS episodes of the cell; the average equals the cell's gain. Grouping the terms by whether the episode ended on the exhaustion path splits the gain exactly into the part carried by exhausted episodes and the part carried by finished ones (Table~\ref{tab:loss}). The two classes are outcomes of the MAS, so the split locates the loss and does not identify its cause. Per-task counts compare the MAS and SAS pass rates on each task over the three seeds. Of the 150 tasks (50 per domain), 33 improve under the MAS for one of the four OpenAI and Google models, 5 for two and none for three or more, while 18 get worse for all four.

\begin{table}[tbp]
\caption{Where the loss arises, per cell meeting the mechanical bars, at $4\times$ (points). The exhausted and finished parts sum to the gain up to rounding; share: part of the loss carried by exhausted episodes, shown where the gain is negative. Tasks: number of the 50 tasks on which the MAS pass rate is above, equal to or below the SAS pass rate.}
\label{tab:loss}
\begin{center}
\scriptsize
\begin{tabular}{@{}llrrrrc@{}}
\toprule
Model & Domain & Gain & Exhausted & Finished & Share & Tasks better / equal / worse \\
\midrule
\texttt{gpt-5-nano} & airline & $+2.7$ & $+2.0$ & $+0.7$ & -- & 13 / 26 / 11 \\
 & retail & $-23.3$ & $-6.2$ & $-17.1$ & 27\% & 2 / 23 / 25 \\
 & telecom & $-13.3$ & $-3.6$ & $-9.8$ & 27\% & 3 / 28 / 19 \\
\texttt{gpt-5-mini} & airline & $-16.7$ & $-11.8$ & $-4.9$ & 71\% & 6 / 27 / 17 \\
 & retail & $-28.0$ & $-22.9$ & $-5.1$ & 82\% & 5 / 19 / 26 \\
 & telecom & $-11.3$ & $-8.7$ & $-2.7$ & 76\% & 5 / 25 / 20 \\
\texttt{gemini-3.1-flash-lite} & airline & $-16.0$ & $-11.8$ & $-4.2$ & 74\% & 0 / 36 / 14 \\
 & retail & $-36.0$ & $-32.2$ & $-3.8$ & 90\% & 3 / 20 / 27 \\
 & telecom & $-22.0$ & $-13.1$ & $-8.9$ & 60\% & 4 / 23 / 23 \\
\texttt{gemini-3-flash} & airline & $-25.3$ & $-17.6$ & $-7.8$ & 69\% & 2 / 30 / 18 \\
 & retail & $-54.7$ & $-48.7$ & $-6.0$ & 89\% & 0 / 15 / 35 \\
 & telecom & $-80.7$ & $-60.7$ & $-20.0$ & 75\% & 0 / 1 / 49 \\
\texttt{qwen3.5-4b} & telecom & $-71.3$ & $-56.2$ & $-15.1$ & 79\% & 0 / 4 / 46 \\
\midrule
Mean (total for tasks) & & $-30.5$ & $-22.4$ & $-8.1$ & 74\% & 43 / 277 / 330 \\
\bottomrule
\end{tabular}
\end{center}
\end{table}

\subsection{Does the principle pick the right architecture?}
\label{sec:transfer}

\begin{table}[t]
\caption{\textbf{Where the principle predicts a MAS benefit, the MAS loses under the ceiling and at best ties without one.} Top: the 45\% rule applied to each of the 13 cells' own SAS \passone. Bottom: the three cells at or below 45\%, with and without a ceiling (uncapped arm). Index: SAS \passone\ on the model's other two domains; tokens: median MAS completion tokens per episode over the SAS median, without a ceiling.}
\label{tab:rule}
\begin{center}
\scriptsize
\setlength{\tabcolsep}{4pt}
\begin{tabular}{@{}lcccc@{}}
\toprule
SAS 95\% CI relative to 45\% & Cells & Principle predicts & Observed & Sign matches \\
\midrule
entirely above & 10 & MAS loses & 10 losses & 10 of 10 \\
entirely below & 2 & MAS gains & 2 losses & 0 of 2 \\
straddles & 1 & (uninformative) & 1 tie & -- \\
\midrule
all informative cells & 12 & & & 10 of 12 (binomial $p = 0.039$) \\
\bottomrule
\end{tabular}

\vspace{6pt}
\begin{tabular}{@{}lccccc@{}}
\toprule
Cell & SAS [95\% CI] & Index & Gain at $4\times$ [95\% CI] & Gain uncapped [95\% CI] & Tokens \\
\midrule
telecom $\times$ \texttt{gemini-3.1-flash-lite} & 34.0 [24.0, 44.7] & 78.0 & $-22.0$ [$-30.7$, $-12.7$] & $+0.7$ [$-8.7$, $+9.3$] & 5.5$\times$ \\
telecom $\times$ \texttt{gpt-5-nano} & 32.7 [22.7, 43.3] & 55.3 & $-13.3$ [$-20.7$, $-6.0$] & $-10.0$ [$-18.7$, $-2.0$] & 2.6$\times$ \\
airline $\times$ \texttt{gpt-5-nano} & 42.7 [31.3, 54.0] & 50.3 & $+2.7$ [$-6.7$, $+12.7$] & $+1.3$ [$-7.3$, $+10.0$] & 2.7$\times$ \\
\bottomrule
\end{tabular}
\end{center}
\end{table}

Figure~\ref{fig:main}a and Table~\ref{tab:rule} test the principle on the axis it was stated on, each cell's own SAS \passone. No slope is drawn on this axis (Lemma~\ref{lem:coupling}). Of the 13 cells, 12 have a SAS CI that does not straddle 45\%, and all 12 have a gain CI excluding zero. The principle predicts the observed sign in 10 of 12, and where it errs matters more than how often. The ten matches are the cells above 45\%, where any statement of the form ``the SAS wins'' would also be right; both cells below 45\%, the only cells on which the principle says something distinctive, lose where it predicts a gain. Sweeping the threshold from 35 to 55\% leaves this pattern unchanged. Removing the ceiling does not rescue the prediction (Table~\ref{tab:rule}, bottom): on telecom $\times$ \texttt{gemini-3.1-flash-lite} it brings the MAS level with the SAS at 5.5 times the SAS's median token spend, and on the two \texttt{gpt-5-nano} cells it changes nothing. Neither this count nor the original 15 of 16 is powered to establish an accuracy, so our primary reading rests on the slope.

\subsection{H1: no detectable capability dependence}
\label{sec:h1}

\begin{figure}[t]
\begin{center}
\includegraphics[width=\linewidth]{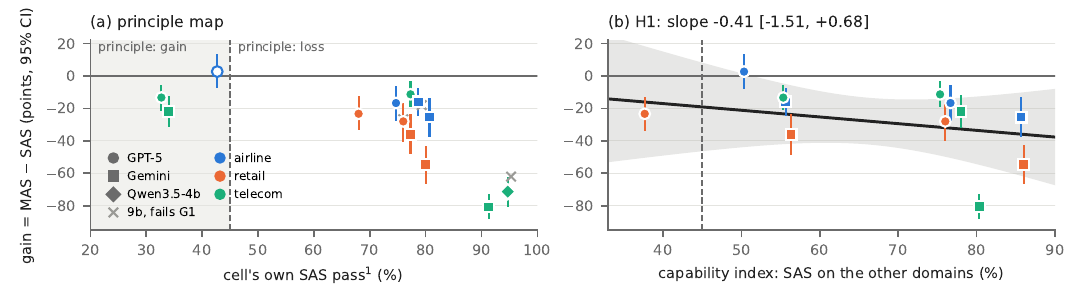}
\end{center}
\caption{(a) Principle map: each of the 13 cells' own SAS \passone\ against its gain with the paired task-cluster 95\% interval; the principle predicts a gain left of the dashed 45\% line and a loss right of it; the hollow marker is the one cell whose interval covers zero; grey crosses are the three \texttt{qwen3.5-9b} cells that fail the mechanical bars. No regression is drawn on this axis. (b) H1: the same gains against the leave-one-domain-out index for the 12 cells of the primary fit, with the weighted fit at the mean family and domain effect and its 95\% band. The two telecom cells left of 45\% in (a) move right of it. Colour marks the domain and shape the family.}
\label{fig:main}
\end{figure}

\begin{table}[t]
\caption{\textbf{The gain shows no detectable dependence on capability, and its fitted level is negative across the observed range.} Left: the H1 slope (points of gain per point of index) under each form; the primary form carries the test (minimum detectable slope in our power analysis $0.20$). $^{a}$Other units. Right: the primary fit at the mean family and domain effect.}
\label{tab:h1}
\begin{center}
\scriptsize
\setlength{\tabcolsep}{4pt}
\begin{minipage}[t]{0.60\linewidth}
\centering
\setlength{\tabcolsep}{3pt}
\begin{tabular}{@{}lccc@{}}
\toprule
Form & Cells & Slope [95\% CI] & Perm.\ $p$ \\
\midrule
leave-one-domain-out index (primary) & 12 & $-0.41$ [$-1.51$, $+0.68$] & 0.43 \\
split seeds & 12 & $-0.44$ [$-0.95$, $+0.07$] & -- \\
logit gain$^{a}$ & 12 & $-1.61$ [$-4.96$, $+1.75$] & -- \\
relative gain$^{a}$ & 12 & $-0.35$ [$-1.40$, $+0.70$] & -- \\
primary with the \texttt{qwen3.5-9b} cells & 15 & $-0.28$ [$-1.26$, $+0.69$] & 0.59 \\
own SAS score (naive, biased) & 12 & $-0.55$ [$-1.16$, $+0.06$] & -- \\
\bottomrule
\end{tabular}
\end{minipage}\hfill
\begin{minipage}[t]{0.37\linewidth}
\centering
\begin{tabular}{@{}lc@{}}
\toprule
Index & Fitted gain [95\% CI] \\
\midrule
37.7\% (lowest cell) & $-16$ [$-50$, $+18$] \\
45\% & $-19$ [$-46$, $+8$] \\
60\% & $-25$ [$-41$, $-9$] \\
86\% (highest cell) & $-36$ [$-62$, $-10$] \\
\midrule
weighted mean, 12 cells & $-28$ [$-44$, $-12$] \\
\bottomrule
\end{tabular}
\end{minipage}
\end{center}
\end{table}

Table~\ref{tab:h1} and Figure~\ref{fig:main}b give the primary test. On the 12 cells with a leave-one-domain-out index (two families, two tiers, three domains; 7 residual degrees of freedom) the slope of gain on the index is $-0.41$ (95\% CI $[-1.51, +0.68]$; permutation $p = 0.43$ two-sided, $0.21$ one-sided), so H1 is not supported and, by our reading rule, the fitted line has no crossing. Every other form agrees in sign and detects no slope, including the naive form, which is biased toward one. No family-by-index interaction is detected (CI $[-3.07, +1.58]$).

This null cannot be read as $|\beta| < 0.20$, the bound our power analysis targeted. The cells scatter around the line far more than task sampling allows: the weighted residual variance is 18.6 times its sampling expectation. What the data do fix is the level: the fitted gain is negative across the observed index range and its interval excludes zero from 60\% upward. The index also relocates the two cells the principle is distinctive about (Table~\ref{tab:rule}, index 78.0 and 55.3\%), so their low scores are a domain effect. The one cell whose index falls below 45\%, retail $\times$ \texttt{gpt-5-nano} (37.7\%), loses 23.3 points.

\subsection{H2: what coordination changes about failure}
\label{sec:h4}

\begin{table}[t]
\caption{\textbf{The MAS does not detectably reduce wrongful writes and converts other failures into exhaustion and inaction.} Pooled over the 13 cells (1,950 episodes per system); paired change per episode in points, cell-cluster bootstrap CI; classes assigned in the order WW, BX, IN, other. FP (airline only; 600 episodes per system) counts judge flags among passes; ``all flags'' adds omission flags, which were not adjudicated by hand, and is an upper bound.}
\label{tab:h4}
\begin{center}
\scriptsize
\setlength{\tabcolsep}{4pt}
\begin{tabular}{@{}lcccc@{}}
\toprule
Outcome & SAS & MAS & Change per episode [95\% CI] & Cells down / up \\
\midrule
pass & 1,362 & 768 & $-30.5$ [$-43.6$, $-18.6$] & 12 / 1 \\
wrongful write (WW) & 55 & 43 & $-0.6$ [$-2.6$, $+1.2$] & 5 / 5 \\
budget exhausted (BX) & 0 & 666 & $+34.2$ [$+24.5$, $+44.3$] & 0 / 13 \\
inaction (IN) & 126 & 268 & $+7.3$ [$+3.7$, $+11.8$] & 1 / 10 \\
other failure & 407 & 205 & $-10.4$ [$-17.0$, $-4.3$] & 10 / 3 \\
\midrule
FP, commission (airline) & 27 of 415 passes & 17 of 332 passes & $-1.7$ (4 clusters, no CI) & 4 / 0 \\
FP, all flags (airline) & 58 of 415 passes & 57 of 332 passes & $-0.2$ (4 clusters, no CI) & 2 / 2 \\
\bottomrule
\end{tabular}
\end{center}
\end{table}

A wrongful write is possible only where the gold trajectory has no write: 24 of 50 tasks on airline, 16 on telecom, 3 on retail. H2a fails on its wrongful-write half: across the 12 SAS cells WW/failures does not decline with the index (slope $-0.15$, 95\% CI $[-0.64, +0.34]$, permutation $p = 0.78$); the rate is high only for \texttt{gpt-5-nano} on airline (33\% of failures). Under the reading grid for H2b the MAS shows no detectable change in wrongful writes (Table~\ref{tab:h4}), so we find no evidence that the architecture acts on the failures it targets. The failures it did change lie elsewhere: exhaustion and inaction rise, and other failures fall. The one cell where the MAS halves wrongful writes, airline $\times$ \texttt{gpt-5-nano} (28 to 14), is the tie of the grid, and there inaction and exhaustion rise by 35 episodes. On airline, the judge flags a commission, an action or offer the policy forbids, in 6.5\% of SAS passes and 5.1\% of MAS passes; about half of the pooled drop follows from the 83 fewer passes of the MAS, and with omission flags included the two systems are level. Commission flags fall with capability (15.6 to 2.5\% of passes), against H2a's second half, on four cells of one domain.

\paragraph{Counts per cell.} Applying the taxonomy of \S\ref{sec:measures} to every episode of the 13 cells, with the database agent's executed calls counted as the MAS's writes, reproduces our earlier count for the gate run (2 wrongful writes among 29 failures); Table~\ref{tab:taxonomy} gives the counts per cell.

\paragraph{FP per cell.} \texttt{nl\_screen.py} ran the benchmark's own assertion judge (\texttt{gpt-4.1}, temperature 0) on every passing episode of the four airline cells, \Sone\ and \Mone, on 26 September; the gate run's earlier screen supplies airline $\times$ \texttt{gemini-3-flash} \Sone. Of 31 distinct unmet assertions, those whose text begins ``Agent does not / should not / must not / should never / never / should refuse'' are commission-type and count as FP as-is; the rest (duties to detect, verify, identify or communicate) are omission-type and enter only after hand adjudication, which has not been done, so the all-flags count is an upper bound. Per cell, commission / all flags among passes: \texttt{gpt-5-nano} 10 / 20 of 64 (\Sone) and 9 / 22 of 68 (\Mone); \texttt{gemini-3.1-flash-lite} 6 / 12 of 118 and 1 / 10 of 94; \texttt{gpt-5-mini} 8 / 14 of 112 and 6 / 17 of 87; \texttt{gemini-3-flash} 3 / 12 of 121 and 1 / 8 of 83. Pooled change per episode, commission $-1.7$ points and all flags $-0.2$ points; with four clusters no interval is reliable, and part of the commission drop reflects the MAS's fewer passes.

\begin{table}[tbp]
\caption{Failure taxonomy per cell meeting the mechanical bars (episodes out of 150): passes, wrongful writes, budget exhausted, inaction, other.}
\label{tab:taxonomy}
\begin{center}
\scriptsize
\setlength{\tabcolsep}{4pt}
\begin{tabular}{@{}llrrrrrcrrrrr@{}}
\toprule
 & & \multicolumn{5}{c}{\Sone} & & \multicolumn{5}{c}{\Mone} \\
\cmidrule{3-7}\cmidrule{9-13}
Model & Domain & PASS & WW & BX & IN & OTHER & & PASS & WW & BX & IN & OTHER \\
\midrule
\texttt{gpt-5-nano} & airline & 64 & 28 & 0 & 10 & 48 & & 68 & 14 & 23 & 22 & 23 \\
 & retail & 102 & 0 & 0 & 16 & 32 & & 67 & 0 & 22 & 23 & 38 \\
 & telecom & 49 & 12 & 0 & 8 & 81 & & 29 & 5 & 19 & 51 & 46 \\
\texttt{gpt-5-mini} & airline & 112 & 5 & 0 & 12 & 21 & & 87 & 0 & 31 & 24 & 8 \\
 & retail & 114 & 0 & 0 & 8 & 28 & & 72 & 0 & 51 & 17 & 10 \\
 & telecom & 116 & 1 & 0 & 10 & 23 & & 99 & 2 & 24 & 10 & 15 \\
\texttt{gemini-3.1-flash-lite} & airline & 118 & 0 & 0 & 13 & 19 & & 94 & 1 & 39 & 12 & 4 \\
 & retail & 116 & 1 & 0 & 10 & 23 & & 62 & 0 & 72 & 10 & 6 \\
 & telecom & 51 & 0 & 0 & 18 & 81 & & 18 & 5 & 62 & 42 & 23 \\
\texttt{gemini-3-flash} & airline & 121 & 2 & 0 & 11 & 16 & & 83 & 0 & 48 & 17 & 2 \\
 & retail & 120 & 0 & 0 & 8 & 22 & & 38 & 0 & 94 & 16 & 2 \\
 & telecom & 137 & 2 & 0 & 2 & 9 & & 16 & 11 & 95 & 17 & 11 \\
\texttt{qwen3.5-4b} & telecom & 142 & 4 & 0 & 0 & 4 & & 35 & 5 & 86 & 7 & 17 \\
\midrule
Total & & 1362 & 55 & 0 & 126 & 407 & & 768 & 43 & 666 & 268 & 205 \\
\bottomrule
\end{tabular}
\end{center}
\end{table}

\subsection{Robustness, sensitivity and deployment cost}
\label{sec:robust}

R1 re-ran both \texttt{gemini-3.1-flash-lite} cells at $1\times$ with the SAS capped too. Capping the SAS at its own 75th percentile costs it 4.0 and 6.7 points (telecom, airline), which sizes the asymmetry of \S\ref{sec:budget}. The gain is negative at both budgets ($-27.3$ and $-24.0$ at $1\times$, $-22.0$ and $-16.0$ at $4\times$), and its change between them has an interval covering zero on both cells (Appendix~\ref{app:card}), so no budget dependence is claimed; the larger budget cuts exhausted MAS episodes from 98\% to 45 and 31\% and lifts the MAS by 9 to 15 points, without closing the gap. R2 re-ran three airline cells with the benchmark's default simulator, \texttt{gpt-4.1}. On the two Gemini cells the gain barely moves (\texttt{gemini-3.1-flash-lite} $-19.3$ against $-16.0$, change $-3.3$, CI $[-11.3, +6.0]$; \texttt{gemini-3-flash} $-29.3$ against $-25.3$, change $-4.0$, CI $[-10.7, +2.7]$). On \texttt{gpt-5-nano}, the straddling cell, the SAS falls to 37.3\% (CI $[30.0, 45.3]$) and the MAS gains $+10.0$ (CI $[+1.3, +19.3]$), the only gain interval above zero in the study; the change between simulators, $+7.3$ (CI $[-3.3, +18.0]$), covers zero, and it is one contrast among many with no multiplicity adjustment. In dollars the MAS costs 1.4 to 3.1 times the SAS per cell (2.3 to 4.0 times for the agent calls alone), and the SAS is more accurate and cheaper in 11 of the 12 priced cells, all but the straddling cell (Table~\ref{tab:costs}). The strong SAS, which separates a loss from dividing the work from a loss from the scaffolding, is reported next (\S\ref{sec:extension}).

\subsection{Extension: the strong SAS}
\label{sec:extension}
\label{app:extension}

S2, the strong SAS: one agent, one model, one context; every call sees the conversation as the SAS does under the base prompt plus one role block. Each turn is two calls. \emph{Plan} (tools shown with \texttt{tool\_choice} ``none'') returns a JSON update of a task ledger and a progress ledger, a policy check (the applicable policy lines quoted and what they require) and a draft of the next action; \emph{act} (native tools) reviews the draft against the policy check and the conversation, revises it if needed, and acts. Budget identical to the MAS cell of the same model and domain ($4\Bx$, $c = \Bx$, $w$, $a_{\min}$, the same status line and notice); after the wrap-up notice a turn is one call without tools and without a plan, as in the SAS; an unparsable plan is retried once; an act reply that is JSON text is never sent (a message drafted inside it is used, else the act call is repeated once with a reminder). Comparators: the SAS is the calibration run, the MAS the $4\times$ grid cell, both reused. Outcomes per cell: $\Mone - \text{S2}$ and $\text{S2} - \Sone$ with task-cluster CIs, completion tokens and the S2/\Sone\ ratio, exhaustion and utilisation, review outcomes (kept, revised, replaced), and the mechanical bars. Reading: (1) $\text{S2} - \Sone$ with a CI below zero on a cell means plan-and-review scaffolding inside one context also loses at the same ceiling, so the grid loss there is not specific to dividing the work among agents; (2) $\text{S2} - \Sone$ covering zero or above it while $\Mone - \text{S2}$ lies below zero attributes the loss to the division; (3) any other pattern is reported as mixed. The first smoke (5 tasks $\times$ 1 trial per cell) passed every mechanical bar; three problems were fixed before the full run, after a second smoke round that also passed every bar (JSON act replies reaching a closing call; plans whose draft was a narrative of the plan where the next action belonged, fixed by one sentence in the block; a logging artefact), with reward read last and playing no part. Cost stop \$75. M2 (Figure~\ref{fig:design}b), in which the policy advisor becomes a checker that can veto an order or a customer message, is outlined but not evaluated here. One confound is stated at the outset: S2's plan call sees the tool definitions and the MAS orchestrator does not, so $\Mone - \text{S2}$ measures the division of work together with schema visibility, and the smoke of the fix (\S\ref{sec:compliance}) shows that schema visibility alone moves the share of write orders executed as issued from about 25\% to 72\%.

\paragraph{Results.} Telecom $\times$ \texttt{gemini-3.1-flash-lite} (150 episodes, no infrastructure errors, \$14.04): S2 36.0\% $[28.8, 43.9]$ against 34.0\% for the SAS and 12.0\% for the MAS at $4\times$; per task against the SAS, better on 15, equal on 23 and worse on 12. S2 exhausted 39 of 150 episodes (the MAS 67); its finished episodes passed 54 of 111 and its exhausted ones none. Its median completion tokens per episode were 5.4 times the SAS's (the MAS 5.8). The mechanical bars pass: 33 of 3,664 calls were cut at $c$, all act calls followed by a closing call, and no JSON or cut text reached the customer. Of 39 act replies that came back as JSON, 22 carried a usable message, 38 were repeated once, and 6 repeats that were JSON again became the holding line (6 of 855 customer messages). Over 1,807 turns the review kept the draft 779 times, revised arguments 135 times, changed the tool 419 times (the model drafts tool names that do not exist), reworded the message 306 times and changed the action type 62 times. Reading (2) holds: the loss on this cell comes from dividing the work among agents, or from the schemas the division withheld from the planner. Read with the uncapped arm for the same cell, where the MAS ties the SAS, what the division costs here is budget spent on coordination: under the same ceiling and a similar spend ratio, the single context finishes more episodes and passes more of those it finishes. $\text{S2}-\Sone$ carries the asymmetry of an uncapped SAS comparator.

Airline $\times$ \texttt{gpt-5-nano}, the straddling cell (150 episodes, no infrastructure errors, \$11.30): S2 45.3\% $[37.6, 53.3]$ against 42.7\% for the SAS and 45.3\% for the MAS at $4\times$; per task against the SAS, better on 12, equal on 27 and worse on 11. S2 exhausted 37 of 150 episodes (the MAS 35); its finished episodes passed 58 of 113 and its exhausted ones 10 of 37. Its median completion tokens per episode were 4.9 times the SAS's (the MAS 2.9). The mechanical bars pass: 20 of 3,076 calls were cut at $c$, and no JSON or cut text reached the customer; 18 of 692 customer messages were the holding line. Neither reading (1) nor reading (2) holds ($\text{S2}-\Sone$ $+2.7$ $[-5.3, +11.3]$; $\Mone-\text{S2}$ $0.0$ $[-8.7, +8.7]$), so the cell is mixed: the three systems tie.

Telecom $\times$ \texttt{gpt-5-nano}, the other cell below 45\% (150 episodes, no infrastructure errors, \$20.14): S2 27.3\% $[20.8, 35.0]$ against 32.7\% for the SAS and 19.3\% for the MAS at $4\times$; per task against the SAS, better on 7, equal on 29 and worse on 14. S2 exhausted 64 of 150 episodes (the MAS 23); its finished episodes passed 40 of 86 and its exhausted ones 1 of 64. Its median completion tokens per episode were 4.9 times the SAS's (the MAS 2.8). Split by episode ending, each episode scored against the SAS on the same task, exhausted episodes carry $-8.0$ $[-12.0, -4.2]$ points and finished ones $+2.7$ $[-2.7, +8.2]$, so the shortfall of S2 runs through the budget. The mechanical bars pass: 46 of 6,193 calls were cut at $c$, and no JSON or cut text reached the customer; 58 of 903 customer messages (6.4\%) were the holding line. Reading (1) does not hold ($\text{S2}-\Sone$ $-5.3$ $[-12.7, +1.3]$) and reading (2) holds ($\Mone-\text{S2}$ $-8.0$ $[-14.7, -1.3]$). Across the three cells, S2 does not differ detectably from the SAS, and the MAS loses to S2 on both cells below 45\%.

\subsection{Compute and cost}
\label{sec:compute}
\label{app:compute}

Episodes reported, 7,950 in all: the 16 SAS and 16 MAS grid cells (4,800); the October-build retail \texttt{qwen3.5-4b} SAS and MAS (300); three uncapped MAS cells (450); two MAS cells at $r = 25$ (300); R1, a capped SAS and a $1\times$ MAS on two cells (600); R2, SAS and MAS on three airline cells with the second simulator (900); S2 on three cells (450); and a duplicate retail $\times$ \texttt{qwen3.5-9b} MAS run (150). Pilots and smoke runs are not counted. Commercial API spend recorded by the runner (agent plus simulator at list prices): six Gemini MAS cells about \$88 and their \Sone\ cells about \$43; six GPT-5 MAS cells about \$112 and their \Sone\ cells about \$50; the $r = 25$ check \$39; the uncapped arm \$44.13; the two hosted 9b MAS cells \$8.48; the simulator for the local 4b telecom cells \$9.25 and for the October-build retail pair \$4.37; R1 \$17.23; R2 \$53.76; S2 \$45.48 and its smoke \$1.59; about \$516 in all. The judge pass, the seed replay (no simulator) and the collaborator's runs are not included. Table~\ref{tab:costs} gives the recorded cost per cell.

\begin{table}[tbp]
\caption{Recorded cost per cell in USD at list prices (agent plus simulator; tau2's own cost field). Ratio: MAS over SAS; agent only: the same ratio without the simulator. The local 4-billion model has no per-call price, so its cells show the simulator cost only.}
\label{tab:costs}
\begin{center}
\scriptsize
\begin{tabular}{@{}llrrrr@{}}
\toprule
Model & Domain & \Sone & \Mone\ at $4\times$ & Ratio & Agent only \\
\midrule
\texttt{gpt-5-nano} & airline & 4.66 & 9.15 & 2.0 & 2.9 \\
 & retail & 3.74 & 7.49 & 2.0 & 3.1 \\
 & telecom & 8.71 & 12.57 & 1.4 & 2.5 \\
\texttt{gpt-5-mini} & airline & 8.77 & 27.54 & 3.1 & 3.9 \\
 & retail & 6.82 & 20.83 & 3.1 & 4.0 \\
 & telecom & 16.94 & 34.45 & 2.0 & 3.4 \\
\texttt{gemini-3.1-flash-lite} & airline & 4.53 & 8.10 & 1.8 & 2.3 \\
 & retail & 4.21 & 7.37 & 1.8 & 2.3 \\
 & telecom & 5.53 & 9.53 & 1.7 & 2.6 \\
\texttt{gemini-3-flash} & airline & 8.52 & 19.43 & 2.3 & 2.6 \\
 & retail & 6.36 & 14.01 & 2.2 & 2.7 \\
 & telecom & 14.29 & 29.47 & 2.1 & 2.9 \\
\texttt{qwen3.5-4b} (simulator only) & telecom & 5.09 & 4.16 & -- & -- \\
\texttt{qwen3.5-4b} (Oct.\ build, simulator only) & retail & 2.10 & 2.27 & -- & -- \\
\texttt{qwen3.5-9b} & retail & -- & 3.28 & -- & -- \\
 & telecom & -- & 5.20 & -- & -- \\
\bottomrule
\end{tabular}
\end{center}
\end{table}

The 4-billion model ran on one laptop (Apple M4 Pro, 48 GB): on telecom, 150 \Sone\ episodes at about 2.3 minutes each and 150 MAS episodes in about 30 hours at one request at a time; on retail (October build), the \Sone\ took about 47 hours of wall time on a laptop shared with other work and the MAS about 25 hours. The seed replay took 86 minutes of model time.

%% file: sections/05_discussion.tex
\section{Discussion and limitations}
\label{sec:discussion}

\paragraph{What the evidence says.} The capability-saturation principle makes two claims: coordination helps while the SAS is weak, and its value declines as the SAS grows stronger. On the two cells that test the first claim in this setting, the MAS loses by 13 and 22 points; we find no support for the prediction in this setting, for this architecture. The one gain in the study appears on the straddling cell under a second simulator. The second claim is left undecided: the slope test detects no dependence, and its interval is too wide to exclude a steep decline, because model-by-domain effects dominate the variation between cells. The mechanism that motivates a MAS here, catching wrongful writes, shows no detectable effect. For systems that serve users under a written policy, the SAS was the better choice at every capability level we measured under a fixed simulator, and at a fraction of the cost. These results are for the MAS as built. Its planner never saw the tool schemas, and \S\ref{sec:compliance} reports how far this moved its orders away from what the database agent executed. A MAS design for such settings therefore needs compute-matched evidence from the setting itself, since results from autonomous benchmarks do not carry over.

\paragraph{Why conversation with a user penalises coordination.} The user holds information the agent needs, on telecom part of the control as well, and the benchmark is half-duplex, so every tool call passes through the user-facing agent's turn and the orchestrator plans before each assignment. The user supplies information turn by turn, so the task cannot be decomposed in advance, and decomposable tasks are where coordination drew its largest benefits in the original study \citep{kim2026capable}. The reward is a state at the end of a long conversation, so a MAS that spends its ceiling early fails outright.

\paragraph{Limitations.} (1) One architecture is tested, and it is reported as built. Its orchestrator plans without the tool schemas that only the database agent holds, so under the stricter audit of \S\ref{sec:compliance} only 25\% of its write orders were executed as issued, against the 90\% the plan required; the pre-registered fidelity bar fails in every cell and the result is a labelled deviation. A fix that shows the orchestrator the schemas raised the as-issued share from 25\% to 72\% in a three-episode smoke test, but the grid was not re-run with it, so the paper measures the architecture with its planner blind to the schemas. A null below 45\% is stronger evidence against transfer than the original best-of-four validation gives, and a null above 45\% is weaker. (2) Only two cells lie clearly below 45\%, so the principle's distinctive prediction rests on two cells and the slope is identified mostly above the threshold. (3) The retail \texttt{qwen3.5-4b} cell was run on a later model build than the one registered, so it sits outside the fits; it shows a 51-point MAS loss at 74\% SAS accuracy (Appendix~\ref{app:local4b}). The airline \texttt{qwen3.5-4b} MAS was run by a collaborator on the registered build against a budget taken from an S1 on a different software stack; paired with the September S1 it is reported as a sensitivity whose sign depends on how that S1's crashed episodes are scored (Appendix~\ref{app:local4b}). (4) The user is a model, and simulators are more cooperative than people, so where a cell sits relative to 45\% may differ with human customers; swapping the simulator already turns the straddling cell's tie into a 10-point MAS gain (R2). (5) The official reward is blind to verbal violations on airline and telecom; we measure this blind spot on airline only, commission flags are classified by a fixed text rule, and the judge was not checked against human labels. (6) The models postdate the benchmark's release and may have seen it in training, which inflates SAS and MAS scores alike; the 4-billion open-weight model's telecom scores (94.7\% and 95.3\% across builds) sit above every closed model's, which is consistent with contamination though we cannot test it. (7) On the local route the benchmark passes each trial's seed to the model, so one seed value fixed a long-reasoning behaviour in the MAS orchestrator across three local cells; the effect is confirmed by a replay and the cells are reported per trial, with no trial excluded (Appendix~\ref{app:local4b}). Further reporting notes, on the model snapshots, the parser, the meters, the judge, the arms and the collaborator's runs, are collected in Appendix~\ref{app:notes}.

%% file: sections/06_conclusion.tex
\section{Conclusion}
\label{sec:conclusion}

Scaling principles for agent systems have so far been derived without a user. We tested the most prominent of them where agents serve users in conversation, on the dual-control benchmark: a centralized MAS built from the same model, reported as built, showed no benefit over the stock SAS in any of the 13 cells that meet the mechanical validity bars under a fixed simulated user, lost where the principle predicts a benefit, drew level at best without a budget ceiling, and did not detectably reduce the wrongful writes it was designed to catch. For agents that serve users, the burden of proof lies with the multi-agent design. We release every trajectory, ledger, prompt and our code, with the pre-registered plan and its change log (code at \url{https://github.com/ethanyyang/agent-scaling-dual-control}; data at \url{https://doi.org/10.5281/zenodo.23458286}), so that the test can be rerun on the next architecture, benchmark and principle.

%% file: sections/appendix.tex
\section{Tests and results card}
\label{app:card}

Table~\ref{tab:card} collects every test in the paper with the rule used to read it and its result, so that each claim in the main text can be traced to one row.

\begin{table}[H]
\caption{Tests and results. Each row gives a test, the rule used to read it and the result.}
\label{tab:card}
\begin{center}
\scriptsize
\setlength{\tabcolsep}{3pt}
\begin{tabular}{@{}>{\raggedright\arraybackslash}p{1.9cm}>{\raggedright\arraybackslash}p{4.4cm}>{\raggedright\arraybackslash}p{4.2cm}>{\raggedright\arraybackslash}p{5.3cm}@{}}
\toprule
Row & Unit; contrast; test & Reading rule & Result \\
\midrule
H1 slope (primary) & cells meeting the mechanical bars; gain on leave-one-domain-out index, family and domain effects, inverse-variance weights; permutation $p$ & supported if slope $< 0$ with CI excluding zero; null read as $|b| < 0.20$ & $-0.41$ $[-1.51, +0.68]$, $p = 0.43$; not supported; CI far wider than the minimum detectable slope of 0.20 \\
H1 crossing & same fit; $x_0 = -\text{intercept}/\text{slope}$; bootstrap CI & transfers if CI covers 45\%; shifts if not; no crossing if slope CI covers zero & no crossing (fitted gain $< 0$ over the whole index range) \\
Sign match & cells not straddling 45\%; predicted vs observed sign; binomial vs 0.5 & $X/N$ beside 15/16; informative and within-noise cells apart & 10 of 12 informative ($p = 0.039$); right on all 10 above 45\%, wrong on both below \\
H2a (primary) & \Sone\ cells; WW/failures and FP/passes on the index & WW/failures slope $< 0$, FP/passes slope not $< 0$ & WW half: $-0.15$ $[-0.64, +0.34]$, $p = 0.78$; not supported; FP half (airline, 4 cells, descriptive): commission flags per pass fall from 15.6 to 2.5\% \\
H2b (primary) & paired \Mone{}$-$\Sone\ WW, pooled, cluster CI & grid over WW, IN, BX, FP & WW $-0.6$ pts $[-2.6, +1.2]$: no detectable change; BX $+34$, IN $+7$ pts; FP commission 6.5 to 5.1\% of passes \\
R1 & 2 cells, gain across budget levels & dependence only if its CI excludes zero & $\text{gain}_{4\times} - \text{gain}_{1\times}$ $+5.3$ $[-2.7, +14.0]$ and $+8.0$ $[-1.3, +18.0]$: not claimed \\
R2 & 3 airline cells, 2 simulators & simulator effect if the change in gain excludes zero & not detected: change $+7.3$ $[-3.3, +18.0]$ (\texttt{gpt-5-nano}, whose gain is $+10.0$ $[+1.3, +19.3]$ under \texttt{gpt-4.1}), $-3.3$ $[-11.3, +6.0]$ (\texttt{gemini-3.1-flash-lite}) and $-4.0$ $[-10.7, +2.7]$ (\texttt{gemini-3-flash}) \\
G1 & every MAS cell; four mechanical bars and one fidelity bar ($\ge 90\%$ of write orders executed as issued) & all bars met, or protocol failure (fix and re-run) & mechanical bars: 13 of 16 met; three \texttt{qwen3.5-9b} cells fail. Fidelity bar: fails in every cell (9 to 41\% per cell, 25\% pooled, stricter audit); fix built and smoke-tested (25\% to 72\%), grid not re-run; MAS reported as built, labelled deviation \\
Local-route seed check & retail $\times$ \texttt{qwen3.5-4b} (Oct.\ build); 6 cut tasks, 4 controls; trial-2 and trial-1 first plan prompts replayed under seed 361454, seed 373753 and no seed; 120 calls & seed explanation supported if trial-2 prompts cut $\ge 4$ of 6 under 361454 and $\le 1$ of 6 unseeded & 5 of 6 under 361454; 0 of 6 unseeded; 0 of 6 under 373753; trial-1 prompts 1 of 6 under 361454; controls never cut: supported. Cells reported per trial, no trial excluded \\
Local 4B cells & telecom (registered build, in grid); retail (Oct.\ build, outside fits); pooled and per trial & reported as built & telecom $-71.3$ $[-80.0, -62.7]$ (trials 0+1 $-69.0$; trial 2 $-76.0$); retail $-50.7$ $[-60.7, -40.7]$ (trials 0+1 $-46.0$; trial 2 $-61.2$); retail fails two mechanical bars \\
Airline $\times$ \texttt{qwen3.5-4b} (sens.) & collaborator's MAS on the registered build paired post hoc with the collaborator's September SAS on another stack & sensitivity only; not in any grid, fit or claim & MAS 56.0\%; gain $+16.3$ $[+4.7, +27.7]$ with the SAS's 69 crashed episodes scored 0, $-7.0$ $[-20.2, +5.3]$ with them excluded; sign depends on the treatment \\
Cell coverage & \Sone\ CI vs 45\% & $\ge 3$ below and $\ge 3$ above for a balanced test & 2 below, 1 straddling, 13 above \\
Uncapped arm (sec.) & 3 cells, no ceiling & (1) ceiling hid a benefit; (2) it did not & (2) met \\
S2 (ext.) & 3 cells (telecom $\times$ \texttt{gemini-3.1-flash-lite}; telecom $\times$ \texttt{gpt-5-nano}; airline $\times$ \texttt{gpt-5-nano}); strong SAS at $4\times$ against the SAS and the MAS & (1) $\text{S2}-\Sone$ CI $<0$: scaffolding loses too; (2) $\text{S2}-\Sone$ covers zero or above and $\Mone-\text{S2}<0$: loss from the division; (3) mixed & telecom $\times$ \texttt{gemini-3.1-flash-lite}, (2): $\text{S2}-\Sone$ $+2.0$ $[-7.3, +12.0]$; $\Mone-\text{S2}$ $-24.0$ $[-33.3, -15.3]$. Telecom $\times$ \texttt{gpt-5-nano}, (2): $\text{S2}-\Sone$ $-5.3$ $[-12.7, +1.3]$; $\Mone-\text{S2}$ $-8.0$ $[-14.7, -1.3]$. Airline $\times$ \texttt{gpt-5-nano}, (3): $\text{S2}-\Sone$ $+2.7$ $[-5.3, +11.3]$; $\Mone-\text{S2}$ $0.0$ $[-8.7, +8.7]$ \\
\bottomrule
\end{tabular}
\end{center}
\end{table}

\section{The local 4-billion model: builds, per-trial results and the seed check}
\label{app:local4b}

\paragraph{Builds.} The registered 4-billion model is \texttt{qwen3.5:4b} at Q4\_K\_M as served by Ollama in September 2026, pinned under a local name; the telecom SAS and MAS ran on it. Before the retail cell could run, the model library replaced the build (a 4.2-billion-parameter weight file with a separate vision projector in place of one 4.7-billion-parameter file) and the registered build could no longer be pulled; the pinned local name was rebuilt on the new build. The digests of both builds are recorded in the released provenance files. The retail SAS calibration and the retail MAS were run on the October build with every other setting as registered (context 131,072, shipped sampling defaults, one request at a time). Because the model differs from the one in the grid, the cell is labelled, kept out of every fit and secondary analysis, and reported here; its SAS does not enter the capability index.

\paragraph{Retail $\times$ \texttt{qwen3.5-4b}, October build.} The SAS ran 150 episodes; five ended in empty model replies after one re-run and are excluded, leaving 145, with \passone\ 74.0\% (task mean; 109 of 145 episodes) and the calibration of Table~\ref{tab:calibration} ($\Bx = 5{,}369$, $B = 21{,}476$). The MAS ran 150 episodes with no infrastructure error: \passone\ 23.3\% $[17.3, 30.7]$, a paired gain of $-50.7$ $[-60.7, -40.7]$ (task-cluster bootstrap, the SAS restricted to its evaluated episodes); 74 of 150 episodes exhausted the ledger (49\%), median utilisation 0.98, finished episodes passed 33 of 76 and exhausted ones 2 of 74; median completion tokens per episode 21,080 against 3,845 (5.5 times); simulator cost \$2.27 against \$2.10. The cell fails two mechanical bars as well as the fidelity bar (Table~\ref{tab:validity}): 15.6\% of tokens fell in cut calls and five ordinary turns carried two cut calls; 3.6\% of calls were cut, no JSON or cut text reached the customer, 347 of 1,630 plan calls did not parse, 80 customer messages were the holding line, and 13 of 79 write orders were executed as issued. The pattern matches the telecom cell of the same model on the registered build: a large loss at a high SAS level, carried by exhaustion.

\paragraph{Per trial.} Each cell runs three trials on the same 50 tasks. On the local route the benchmark writes the trial's seed into the model call (trial 0: 626729; trial 1: 373753; trial 2: 361454); the hosted APIs of the closed models drop the seed. Table~\ref{tab:pertrial} shows that trial 2 exhausts the ledger far more often than trials 0 and 1 in all three local MAS cells, including the collaborator's airline run on the registered build. The pooled gain of each cell is the registered statistic and stands; the per-trial split is reported beside it, and no trial is excluded. Removing trial 2 would move the telecom gain from $-71.3$ to $-69.0$ and the retail gain from $-50.7$ to $-46.0$, with no change of sign or of any interval's side of zero; on airline the pass rate hardly moves across trials, because airline tasks that need no database change pass under exhaustion. The retail cell's pooled pass rate was seen by us before the replay below was designed; this is recorded in the change log, and the reporting rule (pooled with the per-trial split, no exclusion) was fixed before the replay ran.

\begin{table}[H]
\caption{The three local 4-billion MAS cells per trial: MAS passes and exhausted episodes out of 50, SAS passes out of the evaluated episodes of the same trial, and the paired gain (points, task-cluster bootstrap CI) pooled, on trials 0 and 1, and on trial 2. The airline SAS on the registered build does not exist beyond 10 episodes, so that row has no gain.}
\label{tab:pertrial}
\begin{center}
\scriptsize
\setlength{\tabcolsep}{3pt}
\resizebox{\linewidth}{!}{%
\begin{tabular}{@{}llccclll@{}}
\toprule
Cell & Build & \multicolumn{3}{c}{MAS pass / exhausted (trial 0, 1, 2)} & Gain, pooled & Trials 0 and 1 & Trial 2 \\
\cmidrule{3-5}
 & & 0 & 1 & 2 & & & \\
\midrule
telecom (grid) & registered & 15 / 26 & 11 / 27 & 9 / 41 & $-71.3$ $[-80.0, -62.7]$ & $-69.0$ $[-78.0, -59.0]$ & $-76.0$ $[-88.0, -62.0]$ \\
 & SAS passes & 48 of 50 & 47 of 50 & 47 of 50 & & & \\
retail (appendix) & October & 16 / 19 & 14 / 12 & 5 / 43 & $-50.7$ $[-60.7, -40.7]$ & $-46.0$ $[-58.0, -34.0]$ & $-61.2$ $[-73.5, -46.9]$ \\
 & SAS passes & 40 of 48 & 34 of 48 & 35 of 49 & & & \\
airline (collaborator) & registered & 28 / 15 & 30 / 14 & 26 / 42 & -- & -- & -- \\
\bottomrule
\end{tabular}}
\end{center}
\end{table}

\paragraph{Seed replay.} To separate the seed from the trial-2 prompts, the first MAS plan prompt of trial 2 and of trial 1 for ten retail tasks (the six tasks whose trial-2 episodes were exhausted with cut plan calls, tasks 13, 53, 54, 59, 104 and 108, and four controls, tasks 1, 75, 93 and 101) was replayed on a fresh server of the October build under three conditions, seed 361454, seed 373753 and no seed, twice each: 120 calls, with the prompt-token count matching the original run in all 120. The reading rule was fixed before any call: with $k = 6$ cut tasks, the seed explanation is supported if the trial-2 prompts are cut at the cap on at least 4 of 6 first repeats under seed 361454 and on at most 1 of 6 unseeded. Table~\ref{tab:replay} gives the result: the trial-2 prompts under seed 361454 were cut on 5 of 6 first repeats (over both repeats, median 5,369 completion tokens, the cap, with 8 of 12 cut and 7 of 12 returning no visible text), and never under seed 373753 or without a seed; the trial-1 prompts were cut once under 361454 and otherwise never; the control tasks were never cut under any condition (medians 321 to 644 tokens). The seed explanation is supported. Determinism on the local route is high: 35 of 40 seeded repeat pairs were byte-identical and 0 of 20 unseeded pairs. The consequence for the paper is the per-trial reporting above and the limitation stated in \S\ref{sec:discussion}; no registered result changes.

\begin{table}[H]
\caption{Seed replay on the October build (17 October 2026; 120 calls, 86 minutes of model time, no errors). For each condition: trial-2 or trial-1 first plan prompt of the six cut tasks; first-repeat cuts at the cap; and over both repeats the number of calls, median completion tokens, calls cut at the cap and calls with no visible text.}
\label{tab:replay}
\begin{center}
\scriptsize
\begin{tabular}{@{}llcrrrr@{}}
\toprule
Prompt & Seed & Cut on first repeat & Calls & Median tokens & Cut & Empty \\
\midrule
trial 2 & 361454 (trial 2's) & 5 of 6 & 12 & 5,369 & 8 & 7 \\
trial 2 & 373753 (trial 1's) & 0 of 6 & 12 & 404 & 1 & 1 \\
trial 2 & none & 0 of 6 & 12 & 629 & 0 & 0 \\
trial 1 & 361454 & 1 of 6 & 12 & 2,798 & 2 & 2 \\
trial 1 & 373753 & 0 of 6 & 12 & 371 & 0 & 0 \\
trial 1 & none & 0 of 6 & 12 & 538 & 0 & 0 \\
\bottomrule
\end{tabular}
\end{center}
\end{table}

\paragraph{Airline $\times$ \texttt{qwen3.5-4b}: a collaborator's MAS, as a sensitivity.} A collaborator ran the airline MAS of the registered build (150 episodes, 7 to 12 October 2026) under a budget of $B = 27{,}329$ and $c = 6{,}832$, calibrated on the collaborator's own September SAS of the model rather than on a SAS from our stack. It passed 84 of 150 episodes (56.0\%, $[48.0, 63.7]$; per trial 28, 30 and 26 of 50; exhausted 15, 14 and 42) and fails the mechanical bars (12.9\% of tokens in cut calls; one ordinary turn with two cut calls) as well as the fidelity bar (17 of 52 orders as issued). The collaborator started a SAS of the same build on our route (10 episodes, 8 evaluated, 7 passed) and stopped it before the MAS, so no SAS on the registered build exists for this cell. The only SAS of the model on airline is the collaborator's September run (150 episodes, 144 evaluated, 20 to 23 September 2026) on a different stack: route \texttt{ollama\_chat} against an unpinned tag with no digest, context 32,768, temperature 0, and a second call whenever the model returned thinking only (56 such recoveries), on the pinned benchmark commit. Of its 144 evaluated episodes 69 ended in an agent error and score zero; its \passone\ is 39.7\% (task mean; 57 of 144 episodes). After both pass rates were known, we paired the two runs (Table~\ref{tab:airline4b}). Scored as recorded, the MAS gains $+16.3$ points $[+4.7, +27.7]$; with the SAS's crashed episodes excluded (75 episodes on 38 tasks) the MAS loses $-7.0$ $[-20.2, +5.3]$. The MAS used 6.1 times the SAS's median completion tokens per episode (26,884 against 4,428). Because the sign of the gain depends on how the SAS's crashes are treated, and the two runs differ in route, pinning, context length and sampling, the pairing is a sensitivity and enters no grid, fit, secondary analysis or claim. A SAS of the registered build on our route, completed to 150 episodes, would replace it. A further collaborator file, an airline MAS of \texttt{qwen3.5-9b} at a later benchmark commit, fails the mechanical bars (35.7\% of tokens in cut calls, nine ordinary double-cut turns) and is not a cell.

\begin{table}[H]
\caption{Airline $\times$ \texttt{qwen3.5-4b} sensitivity: the collaborator's MAS (registered build, 150 episodes) paired with the collaborator's September SAS (other stack, 144 evaluated episodes). Task-mean \passone; paired gain with task-cluster bootstrap CI (seed 300, 2,000 draws); tasks on which the MAS pass rate is above, equal to or below the SAS's.}
\label{tab:airline4b}
\begin{center}
\scriptsize
\begin{tabular}{@{}lrrrlc@{}}
\toprule
Pairing & Tasks & MAS & SAS & Gain [95\% CI] & Tasks better / equal / worse \\
\midrule
SAS as recorded (crashes score 0) & 50 & 56.0\% & 39.7\% & $+16.3$ $[+4.7, +27.7]$ & 20 / 24 / 6 \\
SAS crashed episodes excluded & 38 & 63.2\% & 70.2\% & $-7.0$ $[-20.2, +5.3]$ & 5 / 26 / 7 \\
By trial, SAS as recorded: trial 0 & 49 & 57.1\% & 38.8\% & $+18.4$ $[+2.0, +34.7]$ & 13 / 32 / 4 \\
\quad trial 1 & 48 & 58.3\% & 41.7\% & $+16.7$ $[0.0, +33.3]$ & 12 / 32 / 4 \\
\quad trial 2 & 47 & 51.1\% & 38.3\% & $+12.8$ $[-2.1, +25.5]$ & 9 / 35 / 3 \\
\bottomrule
\end{tabular}
\end{center}
\end{table}

\section{Further reporting notes}
\label{app:notes}

Points of record that bear on reproduction or on reading single numbers, none of which changes a result.

\begin{enumerate}
\item The OpenAI snapshots are scheduled for retirement on 11 December 2026; after that date the released trajectories are the record.
\item The 56\% plan-parse failure rate of \texttt{qwen3.5-4b} is a property of that model under our parser; a parser that completes a missing closing brace is an obvious post hoc variant and was not run.
\item A write followed by a compensating revert leaves the database unchanged and is invisible to the primary measure; three such candidates in the pilot run are listed for hand adjudication.
\item The reasoning-only sensitivity meter cannot split 1.2\% of gate-run output tokens, whose calls carry no details block; the primary meter still counts them.
\item The judge model is pinned; the secondary measure depends on it.
\item The uncapped arm uses the three cells at or below 45\% and the extension the same three, each with three seeds; in the uncapped arm, excluding the 6 telecom $\times$ \texttt{gpt-5-nano} episodes in which plan text reached the customer moves that cell's gain from $-10.0$ to $-12.0$.
\item The \texttt{qwen3.5-9b} \Sone\ values were computed by a collaborator; the per-episode files needed for the paired intervals of two of its cells are not held by us.
\item The open-weight rungs run on non-standard stacks: a local 4-bit server, and an unpinned hosted route that did not enforce the cap on reasoning tokens.
\item The hosted APIs drop the per-trial seed, so the closed-model cells carry no seed effect of the kind found on the local route; whether the SAS on the local route carries a seed effect of its own was not tested.
\item The fix for the fidelity bar was smoke-tested on 5 tasks in each of three cells; its effect on the pass rate was not measured, and the two misses that were policy refusals by the database agent show that no orchestrator-side change can reach the 90\% bar on its own.
\item The collaborator's September SAS runs of \texttt{qwen3.5-4b} on airline and retail, on a different serving stack, scored 39.7 and 36.3\% with 45 to 48\% of episodes ending in agent errors that our stack did not show, so whether the airline cell would sit below 45\% on our stack is unknown.
\end{enumerate}